%% file: main.tex
\documentclass[10pt,twocolumn,letterpaper]{article}

\usepackage{cvpr}              %

\usepackage{graphicx}
\usepackage{amsmath}
\usepackage{amssymb}
\usepackage{booktabs}
\usepackage{bigstrut}
\usepackage{array}
\usepackage{colortbl}
\usepackage[T1]{fontenc}
\usepackage{wrapfig,lipsum}
\usepackage{booktabs}
\usepackage{bigstrut}
\usepackage{multirow}
\usepackage{placeins}
\usepackage{gensymb}
\usepackage{xcolor}
\usepackage{soul}
\usepackage{makecell}
\usepackage{colortbl}
\makeatletter  %
\@namedef{ver@everyshi.sty}{}
\makeatother
\usepackage{tikz}
\usepackage{yfonts}
\usepackage{graphicx}
\usepackage{import}
\usepackage{appendix}
\usepackage[accsupp]{axessibility}  %
\usepackage{hhline}
\usepackage{stfloats}
\usepackage{setspace}
\usepackage{bigstrut}
\usepackage{atbegshi,picture}   
\usepackage{mathtools}
\usepackage{cite} 
\usepackage[font={footnotesize}]{caption}

\usepackage[pagebackref,breaklinks,colorlinks]{hyperref}

\usepackage[capitalize]{cleveref}
\crefname{section}{Sec.}{Secs.}
\Crefname{section}{Section}{Sections}
\Crefname{table}{Table}{Tables}
\crefname{table}{Tab.}{Tabs.}

\newcommand{\tableborder}[0]{\Xhline{8\arrayrulewidth}}
\newcommand{\R}{\mathbb{R}}  %

\DeclareMathOperator*{\minB}{min}

\newcommand{\link}[1]{{\color{blue}\href{#1}{#1}}}

\newcommand*{\affmark}[1][*]{\textsuperscript{#1}}

\newenvironment{myitem}{\begin{list}{$\bullet$}
{\setlength{\itemsep}{-0pt}
\setlength{\topsep}{0pt}
\setlength{\labelwidth}{5pt}
\setlength{\leftmargin}{10pt}
\setlength{\parsep}{-0pt}
\setlength{\itemsep}{0pt}
\setlength{\partopsep}{0pt}}}%
{\end{list}}

\def\confName{CVPR}
\def\confYear{2023}

\begin{document}

\setlength{\belowdisplayskip}{2pt} 
\setlength{\abovedisplayskip}{2pt} 
\setlength{\belowdisplayshortskip}{2pt} 
\setlength{\abovedisplayshortskip}{2pt} 

\title{BundleSDF: Neural 6-DoF Tracking and 3D Reconstruction of Unknown Objects}

\author{Bowen Wen\affmark[] \quad
Jonathan Tremblay\affmark[] \quad Valts Blukis\affmark[] \quad    Stephen Tyree\affmark[] \quad Thomas Müller\affmark[] \and Alex Evans\affmark[] \quad Dieter Fox\affmark[] \quad Jan Kautz\affmark[] \quad Stan Birchfield\affmark[]\\ 
{\affmark[]NVIDIA}
}

\twocolumn[{
\renewcommand\twocolumn[1][]{#1}%
\maketitle

\begin{center}
\vspace{-0.3in}
    \centering
     \includegraphics[width = 0.95\textwidth]{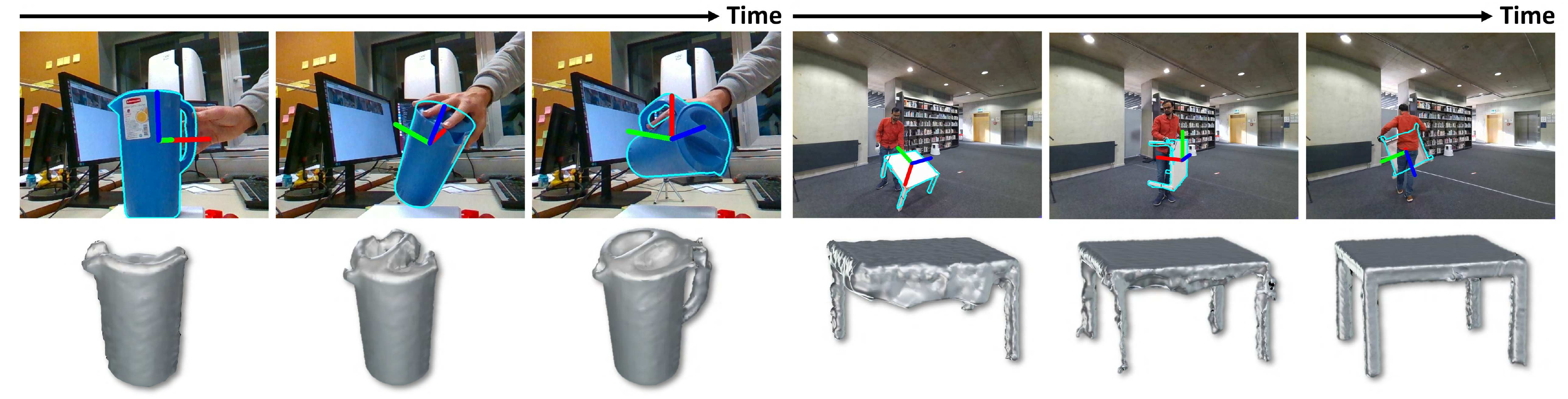}
      \vspace{-0.15in}
    \captionof{figure}{Given a monocular RGBD sequence and 2D object mask (in the first frame only), our method performs causal 6-DoF tracking and 3D reconstruction of an unknown object. Without any prior knowledge of the object or interaction agent, our method generalizes well, handling flat and untextured surfaces, specular highlights, thin structures, severe occlusion, and a variety of interaction agents (human hand / body / robotic arm). The visualized meshes are directly output by the method.}\label{fig:intro}
    \vspace{-.1in}
\end{center}%
}]

\begin{abstract}
\vspace{-0.15in}
We present a near real-time (10Hz) method for 6-DoF tracking of an unknown object from a monocular RGBD video sequence, while simultaneously performing neural 3D reconstruction of the object. Our method works for arbitrary rigid objects, even when visual texture is largely absent. The object is assumed to be segmented in the first frame only.  No additional information is required, and no assumption is made about the interaction agent. Key to our method is a Neural Object Field that is learned concurrently with a pose graph optimization process in order to robustly accumulate information into a consistent 3D representation capturing both geometry and appearance. A dynamic pool of posed memory frames is automatically maintained to facilitate communication between these threads. Our approach handles challenging sequences with large pose changes, partial and full occlusion, untextured surfaces, and specular highlights. We show results on HO3D, YCBInEOAT, and BEHAVE datasets, demonstrating that our method significantly outperforms existing approaches. Project page: \link{https://bundlesdf.github.io/}
\vspace{-0.2in}
\end{abstract}

\section{Introduction}
\vspace{-0.1in}

Two fundamental (and closely related) problems in computer vision are 6-DoF (``degree of freedom'') pose tracking and 3D reconstruction of an unknown object from a monocular RGBD video.
Solving these problems will unlock a wide range of applications in areas such as augmented reality~\cite{marchand2015pose}, robotic manipulation~\cite{kappler2018real,wen2022catgrasp}, learning-from-demonstration~\cite{wen2022you}, and sim-to-real transfer~\cite{andrychowicz2020learning,handa2022dextreme}.

Prior efforts often consider these two problems separately.
For example, neural scene representations have achieved great success in creating high quality 3D object models from real data~\cite{munkberg2022extracting,wang2021neus,oechsle2021unisurf,azinovic2022neural,sun2021neuralrecon,yariv2021volsdf}.  These approaches, however, assume known camera poses and/or ground-truth object masks. Furthermore, capturing a static object by a dynamically moving camera prevents full 3D reconstruction (\eg, the bottom of the object is never seen if resting on a table).
On the other hand, instance-level 6-DoF object pose estimation and tracking methods often require a textured 3D model of the test object beforehand~\cite{li2018deepim,labbe2020cosypose,wang2021gdr,wen2020se,wen2020robust} for pre-training and/or online template matching. While category-level methods enable generalization to new object instances within the same  category~\cite{chen2020learning,tian2020shape,wang2019normalized,li2021leveraging,weng2021captra}, they struggle with out-of-distribution object instances and unseen object categories.

To overcome these limitations, in this paper we propose to solve these two problems jointly.
Our method assumes that the object is rigid, and it requires a 2D object mask in the first frame of the video.
Apart from these two requirements, the object can be moved freely throughout the video, even undergoing severe occlusion.
Our approach is similar in spirit to prior work in object-level SLAM~\cite{Zhu2022CVPR,mccormac2018fusion,sharma2021compositional,merrill2022symmetry,runz2018maskfusion,salasmoreno2013slam,wada2020morefusion}, but we relax many common assumptions, allowing us to handle occlusion, specularity, lack of visual texture and geometric cues, and abrupt object motion.
Key to our method is an online pose graph optimization process, a concurrent Neural Object Field to reconstruct the 3D shape and appearance, and a memory pool to facilitate communication between the two processes.
The robustness of our method is highlighted in Fig.~\ref{fig:intro}.

Our contributions can be summarized as follows: 
\begin{myitem}
    \item A novel method for causal 6-DoF pose tracking and 3D reconstruction of a novel unknown dynamic object.  This method leverages a novel co-design of concurrent tracking and neural reconstruction processes that run online in near real-time while largely reducing tracking drift.
    \item We introduce a hybrid SDF representation to deal with uncertain free space caused by the unique challenges in a dynamic object-centric setting, such as noisy segmentation and external occlusions from interaction.
    \item Experiments on three public benchmarks demonstrate state-of-the-art performance against leading methods.
\end{myitem}

\section{Related Work}

\vspace{0pt}\noindent \textbf{6-DoF Object Pose Estimation and Tracking.} 6-DoF object pose estimation infers the 3D translation and 3D rotation of a target object in the camera's frame. 
State-of-the-art methods often require instance- or category-level object CAD models for offline training or online template matching~\cite{sundermeyer2018implicit,wang2019normalized,labbe2020cosypose,labbemegapose}, which prevents their application to novel unknown objects. Although several recent works \cite{sun2022onepose,liu2022gen6d,park2020latentfusion} relax the assumption and aim to quickly generalize to novel unseen objects, they still require pre-capturing posed reference views of the test object, which is not assumed in our setting. Aside from single-frame pose estimation, 6-DoF object pose tracking leverages temporal information to estimate per-frame object poses throughout the video.
Similar to their single-frame counterparts, these methods make various levels of assumptions, such as training and testing on the same objects~\cite{tjaden2017real,li2018deepim,wen2020se,bundle2021wen,stoiber2022iterative,muller2021seeing} or pretraining on the same category of objects~\cite{lin2022keypoint,wang20206,muller2021seeing}. 
BundleTrack~\cite{bundle2021wen} shares the closest setting to ours, generalizing pose tracking instantly to novel unknown objects. 
Differently, however, our co-design of tracking and reconstruction with a novel neural representation not only results in more robust tracking as validated in experiments (Sec.~\ref{sec:exp}), but also enables an additional shape output, which is not possible with \cite{bundle2021wen}.

\vspace{2pt}\noindent \textbf{Simultaneous Localization and Mapping.}  
 SLAM solves a similar problem to the one addressed in this work, but focuses on tracking the camera pose w.r.t.\ a large static environment \cite{teed2021droid,Zhu2022CVPR,sucar2021imap,mur2015orb}.
Dynamic-SLAM methods usually track  dynamic objects by frame-model Iterative Closest Point (ICP) combined with color \cite{xu2019mid,runz2018maskfusion,ma2015simultaneous,runz2017co}, probabilistic data association \cite{strecke2019fusion}, or 3D level-set likelihood maximization \cite{yuheng2013star3d}. 
Models are simultaneously reconstructed on-the-fly by aggregating the observed RGBD data with the newly tracked pose. In contrast, our method leverages a novel Neural Object Field representation that allows for automatic on-the-fly fusion~\cite{dai2017bundlefusion}, while 
dynamically rectifying historically tracked poses to maintain multi-view consistency. We focus on the object-centric setting including dynamic scenarios, in which there is often a lack of texture or geometric cues, and severe occlusions are frequently introduced by the interaction agent---difficulties that rarely happen in traditional SLAM. Compared to static scenes studied in object-level SLAM~\cite{salasmoreno2013slam,merrill2022symmetry,sharma2021compositional,wada2020morefusion,mccormac2018fusion}, dynamic interaction also allows observing different faces of the object for more complete 3D reconstruction.

\vspace{2pt}\noindent \textbf{Object Reconstruction.}
Retrieving a 3D mesh from images has been extensively studied using learning based 
methods~\cite{lei2020pix2surf,yang2022fvor,munkberg2022extracting}. 
With recent advances in neural scene representation, high quality 3D models can be reconstructed~\cite{munkberg2022extracting,wang2021neus,oechsle2021unisurf,azinovic2022neural,sun2021neuralrecon,yariv2021volsdf}, 
though most of these methods assume known camera poses or ground-truth segmentation and often focus on static scenes with rich texture or geometric cues. In particular, \cite{patten2021object} presents a semi-automatic method with a similar goal but uses manual object pose annotations to retrieve a textured model of the object. In contrast, our method is fully automatic and operates over the video stream causally. Another line of research leverages human hand or body priors to resolve object scale ambiguity or refine object pose estimations via contact/collision constraints~\cite{jiang2022neuralhofusion,zhang2020phosa,xie2022chore,bhatnagar22behave,huang2022intercap,liu2021semi,cao2021reconstructing,yang2021cpf,hasson2020leveraging,krainin2011manipulator}. In contrast, we do not assume specific knowledge of the interaction agent, which allows us to generalize to drastically different forms of interactions and scenarios, ranging from human hand, human body to robot arms, as shown in the experiments. This also eliminates another possible source of error from imperfect human hand/body pose estimation.

\section{Approach}
\vspace{-0.1in}
\begin{figure*}[t]
    \centering
    \includegraphics[width=0.95\textwidth]{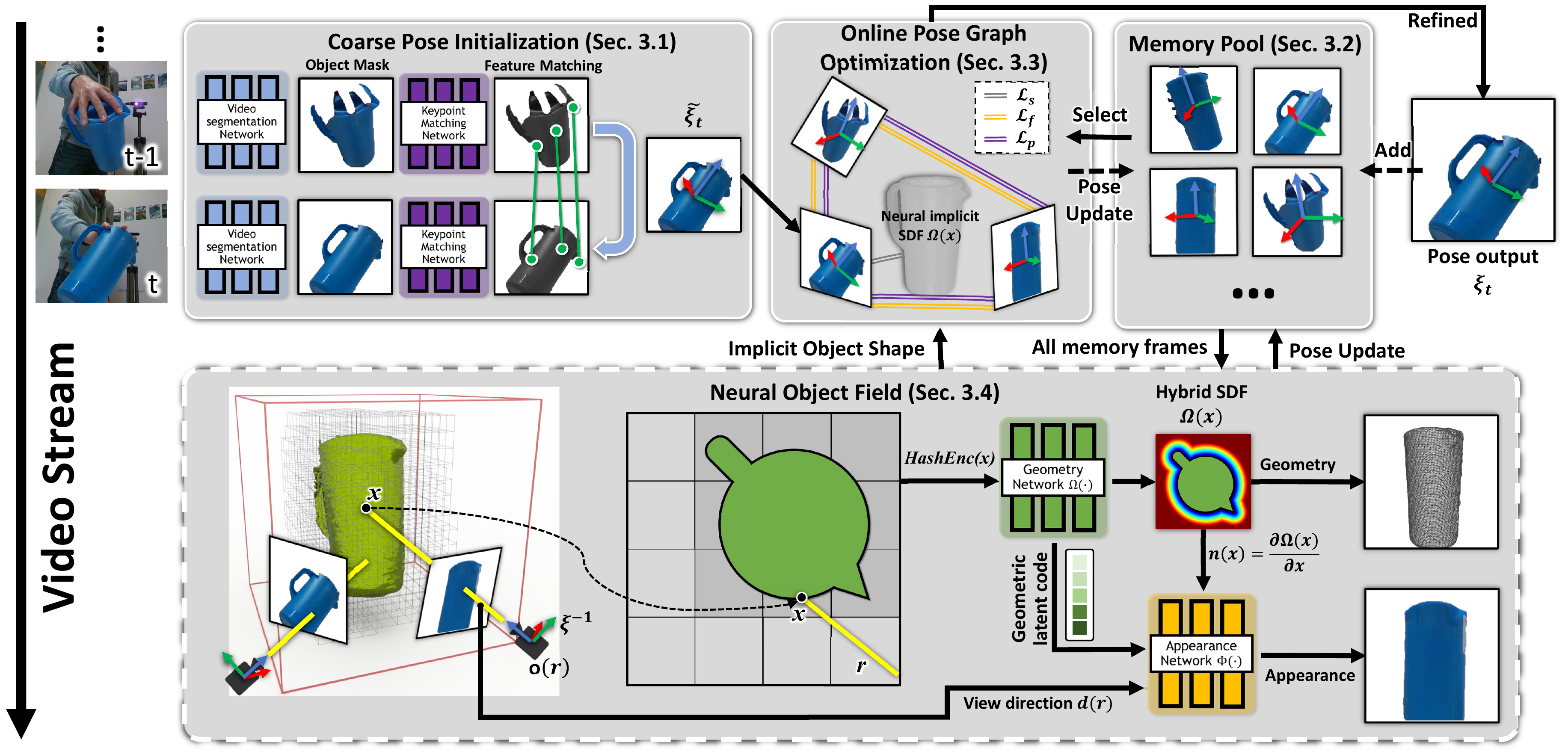}
    \vspace{-0.15in}
    \caption{Framework overview. First, features are matched between consecutive segmented images, to obtain a coarse pose estimate (Sec.~\ref{sec:coarse_pose}).
Some of these posed frames are stored in a memory pool, to be used and refined later (Sec.~\ref{sec:memory_pool}).
A pose graph is dynamically created from a subset of the memory pool (Sec.~\ref{sec:pose_graph}); online optimization refines all the poses in the graph jointly with the current pose.  These updated poses are then stored back in the memory pool.
Finally, all the posed frames in the memory pool are used to learn  a Neural Object Field (in a separate thread) that models both geometry and visual texture (Sec.~\ref{sec:neural_sdf}) of the object, while adjusting their previously estimated poses.}
    \label{fig:pipeline}
    \vspace{-0.25in}
\end{figure*}

An overview of our method is depicted in Fig.~\ref{fig:pipeline}.
Given a monocular RGBD input video, along with a segmentation mask of the object of interest \emph{in the first frame only}, our method tracks the 6-DoF pose of the object through subsequent frames and reconstructs a textured 3D model of the object.
All processing is causal (no access to future frames) The object is assumed to be rigid, but no specific amount of texture is required---our method works well with untextured objects.
In addition, no instance-level CAD model of the object, nor category-level prior (\eg, training on the same object category beforehand), is needed.

\subsection{Coarse Pose Initialization}\label{sec:coarse_pose}
\vspace{-0.1in}
To provide a good initial guess for the subsequent online pose graph optimization, we compute a coarse object pose estimate $\tilde{\xi_t} \in \mathrm{SE}(3)$ between the current frame $\mathcal{F}_t$ and the previous frame $\mathcal{F}_{t-1}$. 
First, the object region is segmented in $\mathcal{F}_t$ by leveraging an object-agnostic video segmentation network~\cite{cheng2022xmem}. 
This segmentation method was chosen because it does not require any knowledge of the object or the interaction agent (\eg, a human hand), thus allowing our framework to be applied to a wide range of scenarios and objects.

Feature correspondences in RGB between $\mathcal{F}_t$ and $\mathcal{F}_{t-1}$ are established via a transformer-based feature matching network~\cite{sun2021loftr}, which was pretrained on a large collection of internet photos~\cite{li2018megadepth}. 
Together with depth, the identified correspondences are  filtered by a RANSAC-based pose estimator~\cite{fischler1981random} using least squares~\cite{arun1987least}. 
The pose hypothesis that maximizes the number of inliers is then selected as the current frame's coarse pose estimation $\tilde{\xi}_t$. 

\subsection{Memory Pool}\label{sec:memory_pool}
\vspace{-0.1in}
To alleviate catastrophic forgetting, which can cause long-term tracking drift, it is important to retain information about past frames.
A common approach exploited by prior work is to fuse each posed observation into an explicit global model~\cite{slavcheva2018sdf,runz2018maskfusion,newcombe2011kinectfusion}. 
The fused global model is then used to compare against the subsequent new frames for their pose estimation (frame-to-model matching).  
However, such an approach is too brittle for the challenging scenarios considered in this work, for at least two reasons.
First, any imperfections in the pose estimates will be accumulated when fusing into the global model, causing additional errors when estimating the pose of subsequent  frames. 
Such errors frequently occur when there is insufficient texture or geometric cues on the object, or this information is not visible in the frame.
Such errors accumulate over time and are irreversible. 
Second, in the case of long-term complete occlusion, large motion changes make registration between the global model and the reappearing frame observation difficult and suboptimal.

Instead, we introduce a keyframe memory pool $\mathcal{P}$ that stores the most informative historical observations. 
To build the memory pool, the first frame $\mathcal{F}_0$ is automatically added, thus setting the canonical coordinate system for the novel unknown object. 
For each new frame, its coarse pose $\tilde{\xi}_t$ is updated by comparing to the existing frames in the memory pool, as described in Sec.~\ref{sec:pose_graph}, to yield an updated pose $\xi_t$.
The frame is only added to $\mathcal{P}$ when its viewpoint (described by $\xi_t$) is deemed to sufficiently enrich the multi-view diversity in the pool while keeping the pool compact.

More specifically, $\xi_t$ is compared with the poses of all existing memory frames in the pool.
Since in-plane object rotation does not provide additional information, this comparison takes into account rotational geodesic distance while ignoring rotation around the camera's optical axis. 
Ignoring this difference allows the system to allocate memory frames more sparsely in the space while maintaining a similar amount of multi-view consistency information. 
This trick enables jointly optimizing a wider range of poses, compared to previous work~(\eg, \cite{bundle2021wen}),
when selecting the same number of memory frames to participate in the online pose graph optimization.

\subsection{Online Pose Graph Optimization} \label{sec:pose_graph}
\vspace{-0.1in}
Given a new frame $\mathcal{F}_t$ with its coarse pose estimation $\tilde{\xi}_t$ (Sec. \ref{sec:coarse_pose}), we select a subset of (no more than) $K$ memory frames from the memory pool to participate in online pose graph optimization. 
The optimized pose corresponding to the new frame becomes the output estimated pose $\xi_t$. 
This step is implemented in CUDA for near real-time processing, making it sufficiently fast to be applied to every new frame, thus resulting in more accurate pose estimations as the object is tracked throughout the video. 

As described below (Sec.~\ref{sec:neural_sdf}), the Neural Object Field is also used to assist in this optimization process.  
Every frame in the memory pool has associated with it a binary flag $b(\mathcal{F})$ indicating whether the pose of this particular frame has had the benefit of being updated by the Neural Object Field.
When a frame is first added to the memory pool, $b(\mathcal{F}) =$~{\sc False}.  This flag remains unchanged through subsequent online updates until the frame's pose has been updated by the Neural Object Field, at which point it is forever set to {\sc True}.

Concurrent with updating the pose of the new frame $\mathcal{F}_t$, all the poses of the subset of frames selected for the online pose graph optimization are also updated to the memory pool, as long as their flag is set to {\sc False}.  
Those frames whose flag is set to {\sc True} continue to be updated by the more reliable Neural Object Field process, but they cease being modified by the online pose graph optimization.

\vspace{0pt}\noindent\textbf{Selecting Subset of Memory Frames}. We constrain the number of memory frames participating in the pose graph optimization to be no more than $K$ for efficiency.
Early in the video, when $|\mathcal{P}|\leqslant K$,  no selection is  needed, and all frames in the memory pool are used. 
When the size of the memory pool grows to be larger than $K$, a selection process is applied with the goal of maximizing the multi-view consistency information. 
Prior efforts select keyframes by exhaustively searching pair-wise feature correspondences and solving a spanning tree~\cite{mur2015orb}, which is either too time-consuming for real-time processing, or simply based on a fixed time interval~\cite{slavcheva2018sdf}, which is less effective in our object-centric setting. 
Therefore, we propose instead to efficiently select the subset $\mathcal{P}_{pg} \subset \mathcal{P}$ of memory frames by leveraging the current frame's coarse pose estimation $\tilde{\xi_t}$ (obtained in Sec.~\ref{sec:coarse_pose}). 
Specifically, for each frame $\mathcal{F}^{(k)}$ in the memory pool, we first compute the point normal map and compute the dot product between these normals and  the ray direction in the new frame's camera view to test their visibility. 
If the point cloud visibility ratio in the new frame $\mathcal{F}_t$ is above a threshold (0.1 for all experiments), we further measure the viewing overlap with $\mathcal{F}_t$ by computing the rotation geodesic distance between $\xi^{(k)}$ and $\tilde{\xi}_t$ while ignoring the in-plane rotation (as described above). 
Finally we select the $K$ memory frames with the maximum viewing overlap (smallest distance) to participate in the pose graph optimization along with $\mathcal{F}_t$.
Therefore, $|\mathcal{P}_{pg}|=K$.

\vspace{0pt}\noindent\textbf{Optimization.} In the pose graph $\mathcal{G}=(\mathcal{V}, \mathcal{E})$, the nodes consist of $\mathcal{F}_t$ and the above selected subset of memory frames:  $\mathcal{V}=\mathcal{F}_t \cup \mathcal{P}_{pg}$, so $|\mathcal{V}|=K+1$.
The objective is to find the optimal poses that minimize the total loss of the pose graph:
\vspace{-0.05in}
\begin{align} 
\mathcal{L}_{pg} = w_{s}\mathcal{L}_s(t)+\!\!\!\!\!\!\!\sum_{i\in \mathcal{V},j\in \mathcal{V},i\neq j}\!\!\!\!\! \left[w_{f}\mathcal{L}_f(i,j) + w_{p}\mathcal{L}_p(i,j)\right], \label{eq:L_pg}
\end{align} 
where $\mathcal{L}_f$ and $\mathcal{L}_p$ are pairwise edge losses~\cite{bundle2021wen}, and $\mathcal{L}_s$ is an additional unary loss.
The scalar factors $w_f, w_p, w_s$ are all set to 1 empirically.
The loss
\begin{align} 
\mathcal{L}_f(i,j)=\sum\limits_{(p_m,p_n) \in C_{i,j}} \rho \left( \left \| \xi_{i}^{-1}p_{m} - \xi_{j}^{-1}p_{n} \right \|_{2} \right)
\end{align} 
measures the Euclidean distance of the RGBD feature correspondences $p_m,p_n \in \R^3$, where $\xi_{i}$ denotes the object pose in frame $\mathcal{F}^{(i)}$, and $\rho$ is the Huber loss~\cite{huber1992robust} for robustness. 
The set of correspondences $C_{i,j}$ between frames $\mathcal{F}^{(i)}$ and $\mathcal{F}^{(j)}$ is detected by the same network introduced in Sec.~\ref{sec:coarse_pose}, where we run batch inference in parallel for efficiency. 
The loss
\begin{align} 
\!\!\!\!\mathcal{L}_p(i,\!j)=\!\!\sum\limits_{p\in I_i}\rho \left( \left|  n_i(p) \cdot \left( T_{ij}^{-1}\pi^{-1}_{D_j} (\pi_j({T_{ij}}       p)) - p \right)    \right| \right)
\end{align} 
measures the pixel-wise point-to-plane distance via re-projective association,
where ${T_{ij}} \equiv \xi_{j}\xi_{i}^{-1}$ transforms from $\mathcal{F}^{(i)}$ to $\mathcal{F}^{(j)}$,
$\pi_j$ denotes the perspective projection mapping onto image $I_j$ associated with $\mathcal{F}^{(j)}$, ${\pi^{-1}_{D_j}}$ represents the inverse projection mapping via looking-up the depth image $D_j$ at the pixel location, $n_i(p)$ denotes the normal via looking-up the normal map of $\mathcal{F}^{(i)}$ at pixel location $p\in I_i$ associated. Lastly, the unary loss
\begin{align} 
\mathcal{L}_s(t)=\sum\limits_{p\in I_t}\rho \big( \left|  \Omega(\xi_t^{-1}(\pi_{D}^{-1}(p))) \right| \big)
\end{align} 
measures the point-wise distance to the neural implicit shape using the current frame, where $\Omega(\cdot)$ denotes the signed distance function from the Neural Object Field as will be discussed in Sec.~\ref{sec:neural_sdf}. 
The Neural Object Field weights are frozen in this step. 
This unary loss is taken into account only after the initial training of the Neural Object Field has converged. 

The poses are represented as inversions of camera poses w.r.t.\ the object, parametrized using Lie Algebra, fixing the coordinate frame of the initial frame as the anchor point. 
We solve the entire pose graph optimization via the Gauss-Newton algorithm with iterative re-weighting. 
The optimized pose corresponding to $\mathcal{F}_t$ becomes its updated pose $\xi_t$. 
For the rest of the selected memory frames, their optimized poses in the memory pool are also updated to rectify possible errors computed earlier in the video, unless $b(\mathcal{F})=$~{\sc True}, as mentioned earlier.

\subsection{Neural Object Field}\label{sec:neural_sdf}
\vspace{-0.1in}

A key to our approach is learning an object-centric neural signed distance field that learns multi-view consistent 3D shape and appearance of the object while adjusting memory frames' poses. It is learned per-video and does not require pre-training in order to generalize to novel unknown objects. This Neural Object Field trains in a separate thread parallel to the online pose tracking. At the start of each training period, the Neural Object Field consumes all the memory frames (along with their poses) from the pool and begins learning. When  training converges, the optimized poses are updated to the memory pool to aid subsequent online pose graph optimization, which fetches these updated memory frame poses each time to alleviate tracking drift. The learned SDF is also updated to the subsequent online pose graph to compute the unary loss $\mathcal{L}_s$ described in Sec.~\ref{sec:pose_graph}. 
The Neural Object Field training process is then repeated by grabbing new memory frames from the pool.

\vspace{0pt}\noindent \textbf{Object Field Representation.} Inspired by 
 \cite{yariv2020multiview}, we represent the object by two functions.  First, the geometry function $\Omega: x \mapsto  s$ takes as input a 3D point $x \in \R^3$ and outputs a signed distance value $s \in \mathbb{R}$.  Second, the appearance function $\Phi: (f_{\Omega(x)}, n, d) \mapsto c$ takes the intermediate feature vector $f_{\Omega(x)} \in \mathbb{R}^3$ from the geometry network, a point normal $n \in \mathbb{R}^3$, and a view direction $d \in \mathbb{R}^3$, and outputs the color  $c \in \mathbb{R}^3_+$. 
In practice, we apply multi-resolution hash encoding \cite{mueller2022instant} to $x$ before forwarding to the network. 
The normal of a point in the object field can be derived by taking the first-order derivative on the signed distance field: $n(x)=\frac{\partial \Omega(x)}{\partial x}$, which we implement by leveraging automatic differentiation in PyTorch~\cite{paszke2019pytorch}. 
For both directions $n$ and $d$, we embed them by a fixed set of low-order spherical harmonic coefficients (order 2 in our case) to prevent over-fitting that could discourage the object pose update (represented as inversion of camera poses w.r.t.\ the object, as mentioned above), in particular the rotations.

The implicit object surface is obtained by taking the zero level set of the signed distance field: 
$S=\left \{ x\in \mathbb{R}^3 \mid \Omega(x)=0  \right \}$. The SDF object representation $\Omega$ has two major benefits compared to \cite{mildenhall2021nerf} in our setting. First, when combined with our efficient ray sampling with depth guided truncation (described below), it enables the training to converge quickly within seconds for online tracking. 
Second, implicit regularization guided by the normals encourages smooth and accurate surface extraction.  This not only provides a satisfactory object shape reconstruction as one of our final goals, but also in return provides more accurate frame-to-model loss $\mathcal{L}_s$ for the online pose graph optimization.

\vspace{0pt}\noindent \textbf{Rendering.} 
Given the object pose $\xi$ of a memory frame, an image is rendered by emitting rays through the pixels.  3D points are sampled at different locations along the ray: 
\begin{align}
    x_i(r) = o(r)+t_i d(r),
\end{align}
where $o(r)$ and $d(r)$ are the ray origin (camera focal point) and ray direction, respectively, both of which depend on $\xi$; and $t_i \in \R_+$ governs the position along the ray. 

The color $c$ of a ray $r$ is integrated by near-surface regions:
\vspace{-0.1in}
\begin{equation}
    c(r)=\int_{z(r)-\lambda}^{z(r)+0.5\lambda} w(x_i)\Phi(f_{\Omega(x_i)},n(x_i),d(x_i))\,dt, \label{eq:render} 
\end{equation}
\begin{equation}
w(x_i)= \frac{1}{1+e^{-\alpha\Omega(x_i)}}\frac{1}{1+e^{\alpha\Omega(x_i)}},
\end{equation}
where $w(x_i)$ is the bell-shaped probability density function \cite{wang2021neus} that depends on the distance from the point to the implicit object surface, \ie, the signed distance $\Omega(x_i)$. $\alpha$ (set to a constant) adjusts the softness of the probability density distribution. The probability reaches a local maximum at the surface intersection. $z(r)$ is the depth value of the ray from the depth image. $\lambda$ is the truncation distance. In Eq.~\eqref{eq:render}, we ignore the contribution from  empty space that is more than $\lambda$ away from the surface to reduce over-fitting from the empty space in the neural field in order to improve pose updates. 
We then only integrate up to a $0.5\lambda$ penetrating distance to model self-occlusion \cite{wang2021neus}. 
An alternative to directly using the depth reading $z(r)$ to guide the integration would be to infer the zero-crossing surface from $\Omega(x_i)$. 
However, we found this requires denser point sampling and slower training convergence compared to using the depth.

\begin{figure}[h]
    \centering
    \definecolor{blue}{RGB}{67, 143, 196}
    \definecolor{orange}{RGB}{255, 192, 0}
    \definecolor{red}{RGB}{255, 0, 0}
    \vspace{-0.15in}
    {\includegraphics[width=0.48\textwidth]{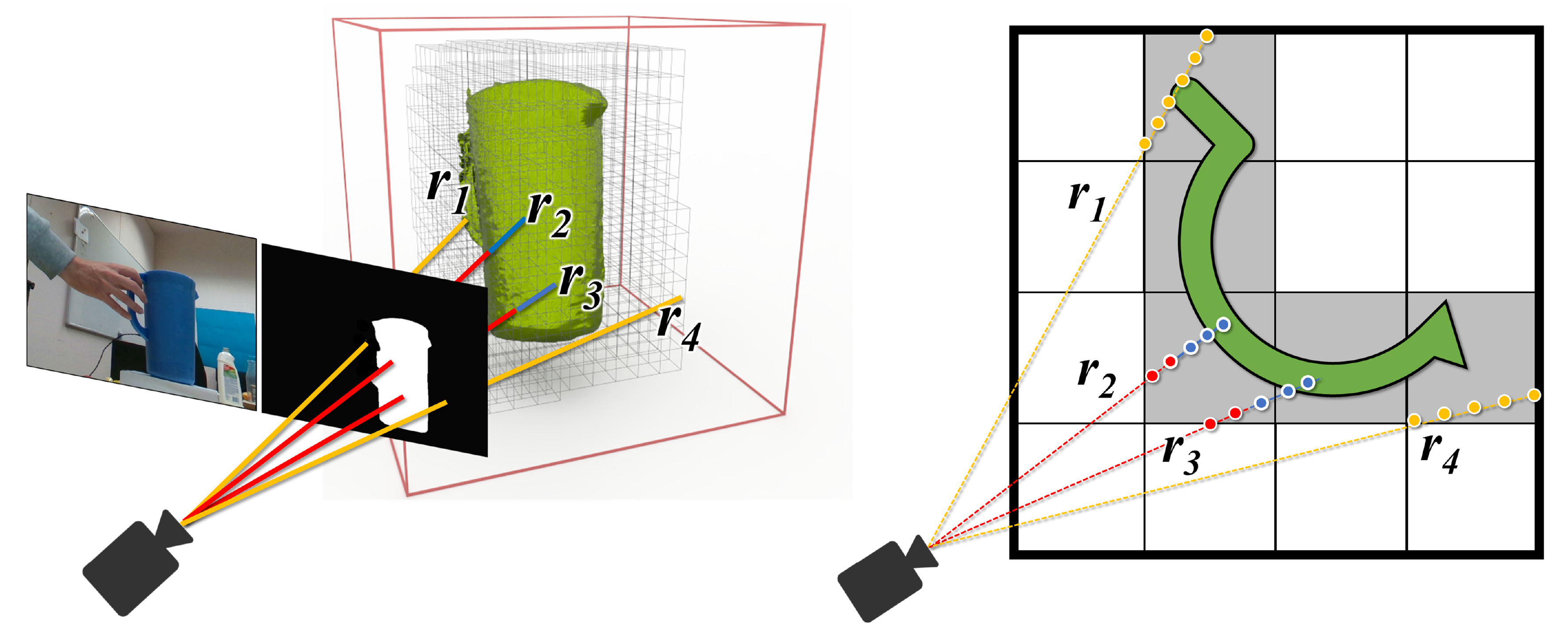}} 
    \vspace{-0.28in}
    \caption{\textbf{Left:} Octree-voxel representation for efficient ray tracing, using the predicted binary mask from the video segmentation network (Sec.~\ref{sec:coarse_pose}), which contains errors.  Rays can land inside the mask (shown as red) or outside (yellow). 
    \textbf{Right:} 2D top-down illustration of the neural volume and point sampling along the rays with hybrid SDF modeling.  Blue samples are near the surface.
    }    \label{fig:ray_sample}
    \vspace{-0.15in}
\end{figure}

\noindent \textbf{Efficient Hierarchical Ray Sampling.} 
For efficient rendering, we construct an Octree representation~\cite{KaolinLibrary} before training by naively merging the point clouds of the posed memory frames. 
We then perform hierarchical sampling along the rays. Specifically, we first uniformly sample $N$ points bounded by the occupancy voxels (gray boxes in Fig.~\ref{fig:ray_sample}), terminating at $z(r)+0.5\lambda$. 
A custom CUDA kernel was implemented to skip the sampling of intermediate unoccupied voxels. 
Additional samples are allocated around the surface for higher quality reconstruction: Instead of importance sampling based on the SDF predictions, which requires multiple forward  passes through the network~\cite{wang2021neus,mildenhall2021nerf}, we draw $N'$ point samples from a normal distribution centered around the depth reading $\mathcal{N}(z(r), \lambda^2)$. This results in $N+N'$ total samples,  without querying the more expensive multi-resolution hash encoding or the networks.

\vspace{0pt}\noindent \textbf{Hybrid SDF Modeling.} Due to the imperfect segmentation and external occlusions, we propose a hybrid signed distance model. Specifically, we divide the space into three regions to learn the SDF (see Fig.~\ref{fig:ray_sample}): 
\begin{myitem}
    \item \textit{Uncertain free space:} These points (yellow in the figure) correspond to the background in the segmentation mask or to pixels with missing depth values, for which the observation is unreliable. For instance, at ray $r_1$'s pixel location in the binary mask, the finger's occlusion results in background prediction, even though it actually corresponds to the pitcher handle. Naively ignoring the background for emitting the ray would lose the contour information, causing bias. Therefore, instead of fully trusting or ignoring \textit{uncertain free space}, we assign a small positive value $\epsilon$ to be potentially external to the object surface so that it can quickly adapt when a more reliable observation is available later:
    \begin{equation}
        \mathcal{L}_{\textit{u}}=\frac{1}{|\mathcal{X}_{\textit{u}}|}\sum_{x\in \mathcal{X}_{\textit{u}}} (\Omega(x)-\epsilon )^2.
    \end{equation}
    \item \textit{Empty space:} These points (red in the figure) are in front of the depth reading up to a truncation distance, making them almost certainly external to the object surface. We apply $L_1$ loss to the truncated signed distance to encourage sparsity:
    \begin{equation}
        \mathcal{L}_{\textit{e}}=\frac{1}{|\mathcal{X}_{\textit{e}}|}\sum_{x\in \mathcal{X}_{\textit{e}}} | \Omega(x)-\lambda |.
    \end{equation}
    \item \textit{Near-surface space:} These points (blue in the figure) are near the surface, no more than $z(r)+0.5\lambda$ distance behind the depth reading to model self-occlusion. This space is critical for learning the sign flipping in SDF and the zero level set. We approximate the near-surface SDF by projective approximation for efficiency:
    \begin{align}
        \mathcal{L}_{\textit{surf}}=\frac{1}{|\mathcal{X}_{\textit{surf}}|}\sum_{x\in \mathcal{X}_{\textit{surf}}}\left(\Omega(x)
        +d_x - d_D \right)^2,
    \end{align}
    where $d_x=\left\|x-o(r)\right\|_2$ and $d_D=\left\|\pi^{-1}(z(r))\right\|_2$ are the distance from ray origin to the sample point and the observed depth point, respectively.
    
\end{myitem}

\noindent \textbf{Training.} The trainable parameters include the multi-resolution hash encoder, $\Omega$, $\Phi$, and the object pose updates in the tangent space parametrized in Lie Algebra $\Delta\overline{\xi}\in \R^{(|\mathcal{P}|-1)\times 6}$, wherein we freeze the first memory frame's pose to be the anchor point. The training loss is:
\begin{equation}
\begin{aligned}
\mathcal{L}=&w_{\textit{u}}\mathcal{L}_{\textit{u}}\!+\!w_{\textit{e}}\mathcal{L}_{\textit{e}}\!+\!w_{\textit{surf}}\mathcal{L}_{\textit{surf}}\!+\!w_{c}\mathcal{L}_{c}\!+\!w_{\textit{eik}}\mathcal{L}_{\textit{eik}},
\end{aligned}
\end{equation}
where $\mathcal{L}_{c}$ denotes the $L_2$ loss over the foreground color for appearance network supervision:
\begin{align}
    \mathcal{L}_{c}=\frac{1}{|\mathcal{R}|}\sum_{r\in \mathcal{R}}\left \| \Phi(f_{\Omega(x)},n(x),d(r))-\bar{c}(r) \right \|_2,
\end{align}
and $\mathcal{L}_{\textit{eik}}$ is the Eikonal regularization \cite{gropp2020implicit} over the SDF in \textit{near-surface space}:
\begin{align}
\mathcal{L}_{\textit{eik}}=\frac{1}{|\mathcal{X}_{\textit{surface}}|}\sum_{x\in \mathcal{X}_{\textit{surface}}} ( \left\|\nabla  \Omega(x)\right\|_2-1 )^2.
\end{align}
Unlike \cite{wang2021neus} which requires ground-truth mask as input, we do not perform mask supervision, since the predicted mask is often noisy from the network.

\vspace{-0.1in}
\section{Experiments}\label{sec:exp}
\vspace{-0.1in}

\subsection{Datasets}
\vspace{-0.1in}
To evaluate our method, we consider three real-world datasets with drastically different forms of interactions and dynamic scenarios. For results on wild application and static scenes, see \href{https://bundlesdf.github.io/}{project page}.

\noindent \textbf{HO3D \cite{hampali2020honnotate}:} This dataset contains the RGBD video of a human hand interacting with YCB objects \cite{calli2015benchmarking}, captured by Intel RealSense camera at close range. Ground truth is automatically generated from multi-view registration. We adopt the most recent version HO-3D\_{v3} and test on the official evaluation set. This results in 4 different objects, 13 video sequences, and 20428 frames in total.

\noindent \textbf{YCBInEOAT \cite{wen2020se}:} This dataset contains the ego-centric RGBD videos of a dual-arm robot manipulating the YCB objects \cite{calli2015benchmarking} captured by Azure Kinect camera at mid range. There are three types of manipulation: (1) single arm pick-and-place, (2) within-hand manipulation, and (3) pick-and-place with handoff between arms. Although this dataset was originally developed to evaluate pose estimation approaches relying on CAD models, we do not provide any object prior knowledge to the evaluated methods. There are 5 different objects, 9 videos, and 7449 frames in total.

\noindent \textbf{BEHAVE \cite{bhatnagar22behave}:} This dataset contains the RGBD video of a human body interacting with the objects, captured at far range by a pre-calibrated multi-view system with Azure Kinect cameras. However, we constrain our evaluation to the single-view setting, where severe occlusions frequently occur. We evaluate on  the official test split excluding the deformable objects. This results in 16 different objects, 70 videos/scenes, and 107982 frames in total.

\subsection{Metrics}
\vspace{-0.1in}
 We separately evaluate pose estimation and shape reconstruction. For 6-DoF object pose, we compute the area under the curve (AUC) percentage of \textit{ADD} and \textit{ADD-S} metrics \cite{xiang2018posecnn,he2022fs6d,bundle2021wen} using ground-truth object geometry. For 3D shape reconstruction, we compute the chamfer distance between the final reconstructed mesh and ground-truth mesh in the canonical coordinate frame defined by the first image of each video. More details can be found in the appendix.

\subsection{Baselines}
\vspace{-0.1in}
We compare against DROID-SLAM (RGBD) \cite{teed2021droid}, NICE-SLAM \cite{Zhu2022CVPR}, KinectFusion \cite{newcombe2011kinectfusion},  BundleTrack \cite{bundle2021wen} and SDF-2-SDF \cite{slavcheva2018sdf} using their open-source implementations with the best tuned parameters.
We additionally include the baseline results from their leaderboard.  Note that methods such as \cite{newcombe2015dynamicfusion,innmann2016volumedeform} focus on deformable objects and the root 6-DoF tracking and fusion are often based on ~\cite{newcombe2011kinectfusion}, whereas we focus on rigid objects that are dynamically moving. We thus omit their comparisons. The inputs to each evaluated method are the RGBD video and the first frame's mask indicating the object of interest. We augment the comparison methods with the same video segmentation masks used in our framework for fair comparison, to focus on 6-DoF object pose tracking and 3D reconstruction performance. In the case of tracking failure, no re-initialization is performed to test long-term tracking robustness.  

DROID-SLAM \cite{teed2021droid}, NICE-SLAM \cite{Zhu2022CVPR} and KinectFusion \cite{newcombe2011kinectfusion} were originally proposed for camera pose tracking and scene reconstruction. When given the segmented images, they run in an object-centric setting. Since DROID-SLAM \cite{teed2021droid} and BundleTrack \cite{bundle2021wen} cannot reconstruct an object mesh, we augment these methods with TSDF Fusion \cite{zeng20163dmatch,curless1996volumetric} for shape reconstruction evaluation. For NICE-SLAM~\cite{Zhu2022CVPR} and our method, we initialize the neural volume's bound   using only the first frame's point cloud (to preserve causal processing, we cannot access future frames).

\begin{figure*}[tbh]
    \centering
    \definecolor{cyan}{RGB}{0, 255, 255}
    \includegraphics[width=0.95\textwidth]{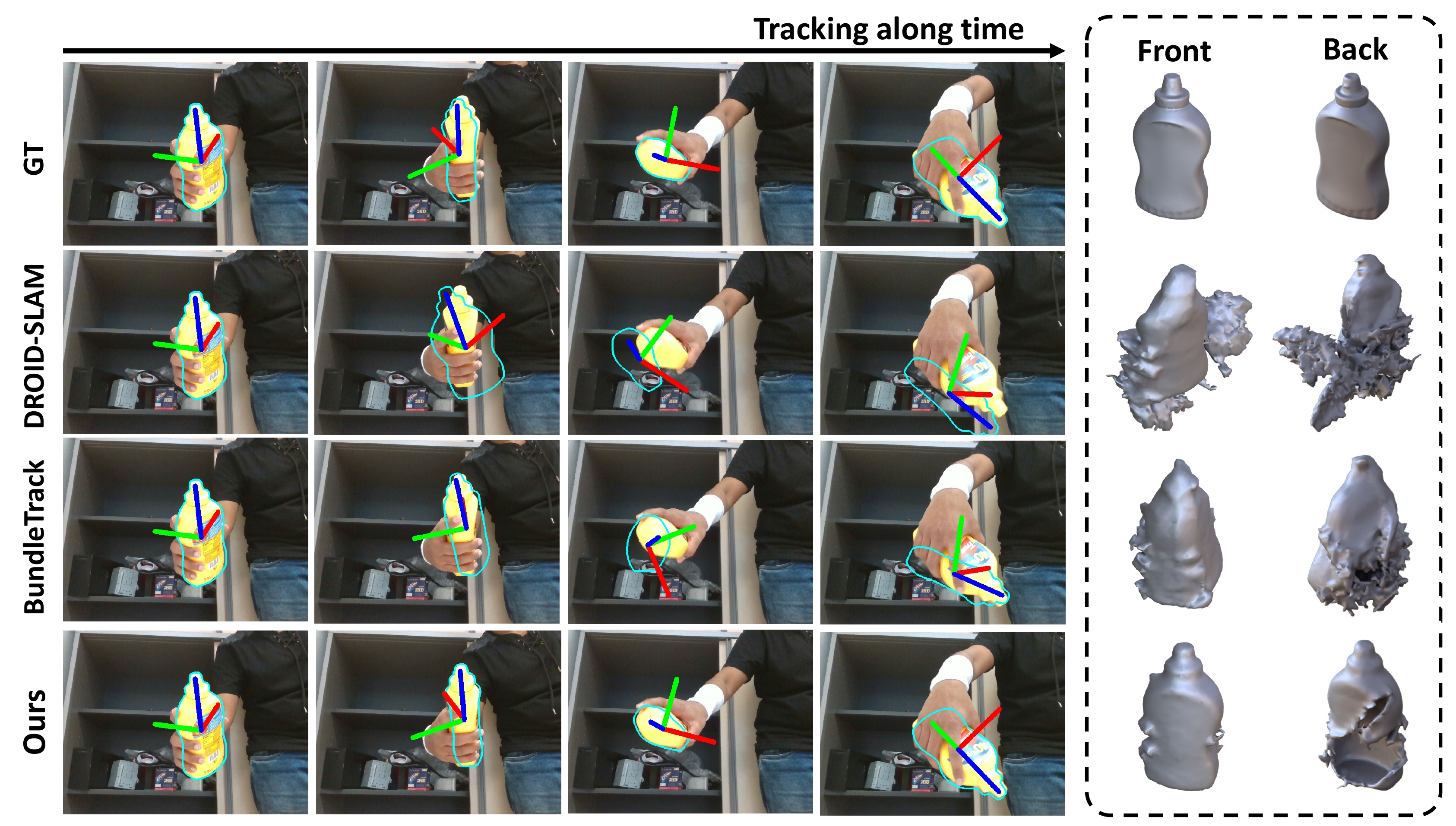}
    \vspace{-0.15in}
    \caption{Qualitative comparison of the three most competitive methods on HO3D Dataset. \textbf{Left:} 6-DoF pose tracking visualization, where the  contour (\textcolor{cyan}{cyan}) is rendered with the estimated pose. Note, as shown in the 2nd column, that our predicted pose sometimes corrects errors in the ground truth. \textbf{Right:} Front and back view of the final reconstructed shape output by each method. Due to hand occlusions, some parts of the object are never visible in the video.  Meshes are rendered from the same viewpoint, though significant drift of DROID-SLAM and BundleTrack results in erroneously rotated meshes.}
    \label{fig:qual}
    \vspace{-0.25in}
\end{figure*}

\subsection{Comparison Results on HO3D}
\vspace{-0.1in}

\begin{figure}[h]
    \centering
    \includegraphics[width=0.47\textwidth]{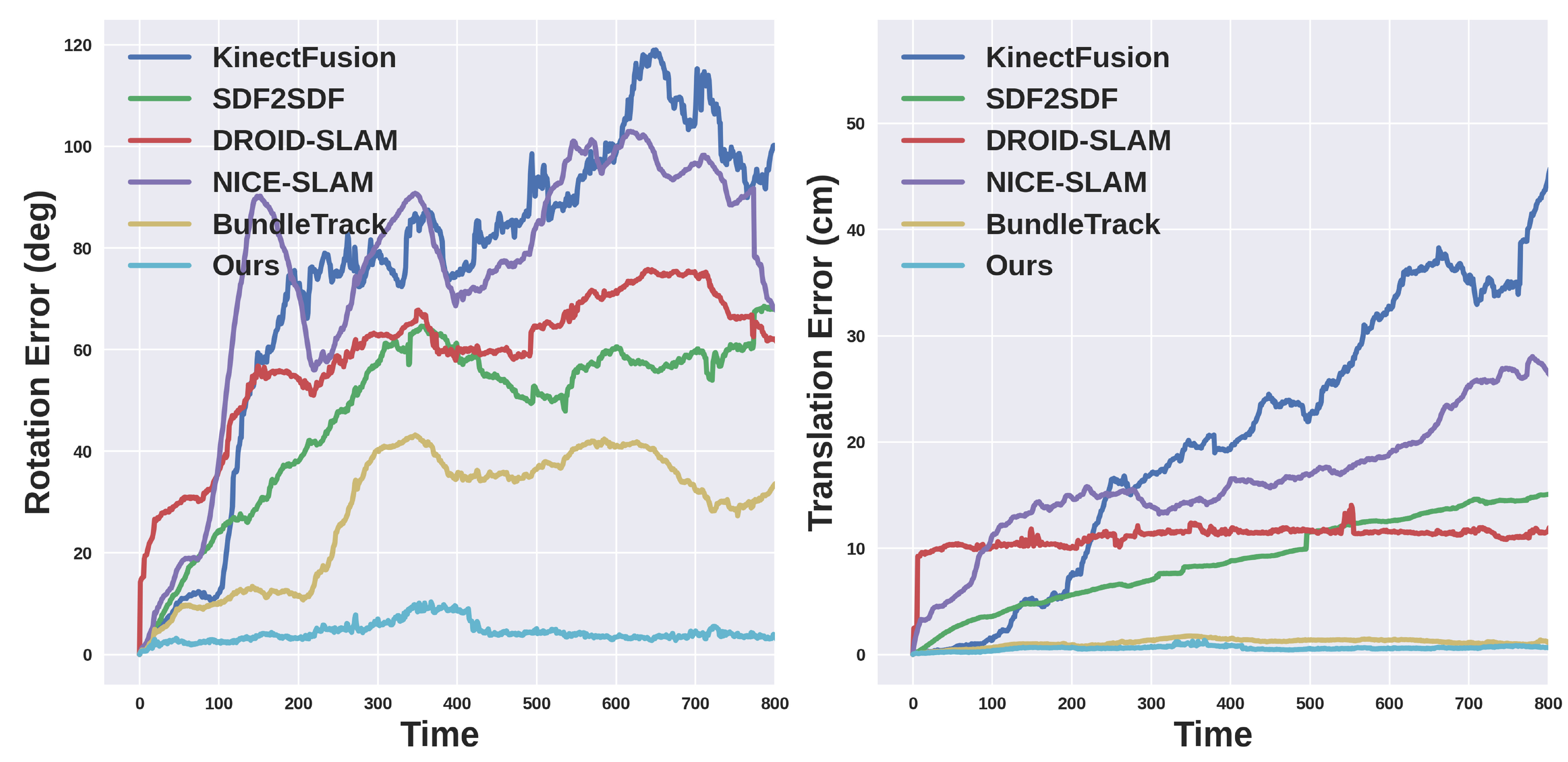}
    \vspace{-0.15in}
    \caption{Pose tracking error against time on HO3D Dataset. Each time stamp's result is averaged across all videos.  \textbf{Left:} Rotation error measured by geodesic distance. \textbf{Right:} Translation error. }
    \label{fig:ho3d_err_time}
    \vspace{-0.2in}
\end{figure}

Quantitative results on HO3D are shown in Tab.~\ref{tab:ho3d} and Fig.~\ref{fig:ho3d_err_time}.  Our method outperforms the comparison methods by a large margin on both 6-DoF pose tracking and 3D reconstruction. For DROID-SLAM \cite{teed2021droid}, NICE-SLAM \cite{Zhu2022CVPR} and KinectFusion \cite{newcombe2011kinectfusion}, when working in an object-centric setting, significantly less texture or geometric (purely planar or cylindrical object surfaces) cues can be leveraged for tracking, leading to poor performance. Fig.~\ref{fig:ho3d_err_time} presents the tracking error against time to study the long-term tracking drift. While BundleTrack \cite{bundle2021wen} achieves similarly low translation error as our approach, it struggles on the rotation estimation.  In contrast, our method maintains a low tracking error throughout the video.  We provide per-video quantitative results in the appendix. 

Fig.~\ref{fig:qual} shows example qualitative results of the three most competitive methods. Despite multiple challenges such as severe hand occlusions, self-occlusions, little texture cues in intermediate observations and strong lighting reflections, our method keeps tracking accurately along the video and obtains dramatically higher quality  3D object reconstruction. Notably, our predicted pose is sometimes more accurate than ground-truth, which was annotated by multi-camera multi-view registration leveraging hand priors.

\input{tables/ho3d.tex}

\vspace{-0.2in}
\subsection{Comparison Results on YCBInEOAT}
\vspace{-0.1in}

\input{tables/ycbineoat.tex}

Quantitative results on YCBInEOAT are shown in Tab.~\ref{tab:ycbineoat}. 
This dataset captures the interaction between the robot arms and the object from an ego-centric view, which leads to challenges due to the constrained camera view and severe occlusions by the robot arms.  For completeness, in this table we also include additional baseline methods from \cite{bundle2021wen}.\footnote{For fair comparison, we only include baselines from \cite{bundle2021wen} that---like our method---do not require instance- or category-level object knowledge.}
The results from these methods, indicated by asterisk ($^*$), are simply copied from \cite{bundle2021wen}.
Note that, in the case of (non-asterisk) BundleTrack, we re-run the algorithm with the same segmentation masks as ours for fair comparison, and we augment with TSDF Fusion for reconstruction evaluation (same as Tab.~\ref{tab:ho3d}).
We omit the re-running for MaskFusion*~\cite{runz2018maskfusion} and TEASER++*~\cite{Yang20troteaser} due to their relatively poorer performance.

Our approach sets a new benchmark record on ADD-S metric and chamfer distance in 3D reconstruction, while obtaining comparable performance with the previous state-art-art method on ADD metric. In particular, while BundleTrack \cite{bundle2021wen} achieves competitive object pose tracking, it does not obtain satisfactory 3D reconstruction results. This demonstrates the benefits of our co-design of tracking and reconstruction.

\vspace{-0.1in}
\subsection{Comparison Results on BEHAVE}

\input{tables/behave.tex}

Quantitative results on BEHAVE are shown in Tab.~\ref{tab:behave}. We refer to the supplemental material for more detailed results. In our setting of single-view and zero-shot transfer without leveraging human body priors, this dataset exhibits extreme challenges. For instance, (i) there are long-term complete occlusions when the human carries the object and faces away from the camera; (ii) severe motion blur and abrupt displacement frequently occur due to the human freely swinging the object; (iii) the objects are of diverse properties and vary greatly in size; (iv) the video is captured at a distance from the camera, making it difficult for depth sensing. Therefore, evaluation on this benchmark pushes the boundary to a more difficult setting. Despite these challenges, our method is still able to perform long-term robust tracking in most scenarios and performs significantly better than previous methods.

\vspace{-0.1in}
\subsection{Ablation Study}

\input{tables/ablations.tex}

We investigate the effectiveness of our design choices on HO3D dataset given its more accurate pose annotations. The results are shown in Tab.~\ref{tab:ablations}. \textit{Ours w/o memory} achieves dramatically worse performance as there is no mechanism to alleviate tracking drift. For \textit{Ours-GPG}, even with similar amount of computation, it struggles on  objects or observations with little texture or geometric cues due to hand-crafted losses. Aside from object pose tracking, \textit{Ours w/o memory}, \textit{Ours w/o NOF} and \textit{Ours-GPG} lack the module for 3D object reconstruction.  \textit{Ours w/o hybrid SDF} ignores the contour information and can be biased by false positive segmentation when rectifying the memory frames' pose. These lead to less stable pose tracking and more noisy final 3D reconstruction. \textit{Ours w/o compact mem pool}, when under the same computational budget, leads to insufficient pose coverage during pose graph optimization and Neural Object Field learning, as mentioned in Sec.~\ref{sec:memory_pool}.

\vspace{-0.1in}
\section{Conclusion}
\vspace{-0.05in}
We presented a novel method for 6-DoF object tracking and 3D reconstruction from a monocular RGBD video.  Our method only requires segmentation of the object in the initial frame. Leveraging two parallel threads that perform online graph pose optimization and Neural Object Field representation respectively, our method is able to handle challenging scenarios, such as fast motion, partial and compete occlusion, lack of texture, and specular highlights.  On several datasets we have demonstrated state-of-the-art results compared with existing methods.  Future work will be aimed at leveraging shape priors to reconstruct unseen parts.

{\small
\bibliographystyle{ieee_fullname}
\bibliography{ref}
}

\clearpage

\appendix
\appendixpage

\input{supplemental}

\end{document}

%% file: tables/ho3d.tex
\begin{table}[h]
\centering
\def\mywidth{0.4\textwidth} 
\resizebox{\mywidth}{!}{
\begin{tabular}{l|rr|r}
\Xhline{8\arrayrulewidth}
       & \multicolumn{2}{c|}{Pose} & \multicolumn{1}{c}{Reconstruction} \\
      & ADD-S (\%) $\uparrow$ & ADD (\%) $\uparrow$ & CD (cm) $\downarrow$ \bigstrut[b]\\
\hline
NICE-SLAM \cite{Zhu2022CVPR} & 22.29 & 8.97  & 52.57 \\
SDF-2-SDF \cite{slavcheva2018sdf} & 35.88 & 16.08 & 9.65 \\
KinectFusion \cite{newcombe2011kinectfusion} & 25.81 & 16.54 & 15.49 \\
DROID-SLAM \cite{teed2021droid} & 64.64 & 33.36 & 30.84 \bigstrut[t]\\
BundleTrack \cite{bundle2021wen} & 92.39 & 66.01 & 52.05 \\
\rowcolor[rgb]{ .847,  .847,  .847} Ours  & \textbf{96.52} & \textbf{92.62} & \textbf{0.57} \\

\Xhline{8\arrayrulewidth}
\end{tabular}%
}
\vspace{-0.1in}
\caption{Comparison on HO3D Dataset. ADD and ADD-S are AUC percentage (0 to 0.1~m).  Reconstruction is measured by chamfer distance.}
\label{tab:ho3d}
\vspace{-0.1in}
\end{table}

%% file: tables/ycbineoat.tex
\begin{table}[h]
\centering
\vspace{-0.in}
\def\mywidth{0.4\textwidth} 
\resizebox{\mywidth}{!}{
\begin{tabular}{l|rr|r}
\tableborder
                               & \multicolumn{2}{c|}{Pose}                                       & \multicolumn{1}{c}{Reconstruction} \bigstrut[t]\\
                               & ADD-S (\%) $\uparrow$          & ADD (\%) $\uparrow$            & CD (cm) $\downarrow$ \bigstrut[b]\\
\hline
NICE-SLAM \cite{Zhu2022CVPR}   & 23.41                          & 12.70                          & 6.13 \\
SDF-2-SDF \cite{slavcheva2018sdf} & 28.20                          & 14.04                          & 2.61 \\
KinectFusion \cite{newcombe2011kinectfusion} & 46.39                          & 34.68                          & 4.63 \\
DROID-SLAM \cite{teed2021droid} & 32.12                          & 20.39                          & 2.34 \\
BundleTrack \cite{bundle2021wen} & 93.01                          & 87.26                          & 2.81 \\
BundleTrack$^*$ \cite{bundle2021wen} & 92.53                          & \textbf{87.34}                 & - \\
MaskFusion$^*$ \cite{runz2018maskfusion} & 41.88                          & 35.07                          & - \bigstrut[t]\\
TEASER++$^*$ \cite{Yang20troteaser} & 81.17                          & 57.91                          & - \\
\rowcolor[rgb]{ .847,  .847,  .847} Ours                           & \textbf{93.77}                 & 86.95                          & \textbf{1.16} \bigstrut[b]\\
\tableborder
\end{tabular}%
}
\vspace{-0.1in}
\caption{Comparison on YCBInEOAT Dataset. ADD and ADD-S are AUC percentage (0 to 0.1~m).  Reconstruction is measured by chamfer distance.} \label{tab:ycbineoat}
\vspace{-0.25in}
\end{table}

%% file: tables/behave.tex
\begin{table}[h]
\vspace{-0.2in}
\centering
\def\mywidth{0.4\textwidth} 
\resizebox{\mywidth}{!}{
\begin{tabular}{l|rr|r}
\Xhline{8\arrayrulewidth}
                               & \multicolumn{2}{c|}{Pose}                                       & \multicolumn{1}{c}{Reconstruction} \bigstrut[t]\\
                               & ADD-S (\%) $\uparrow$          & ADD (\%) $\uparrow$            & CD (cm) $\downarrow$ \bigstrut[b]\\
\hline
DROID-SLAM \cite{teed2021droid} & 56.14                          & 32.29                          & 11.24 \bigstrut[t]\\
BundleTrack \cite{bundle2021wen} & 59.06                          & 45.03                          & 19.27 \\
KinectFusion \cite{newcombe2011kinectfusion} & 38.37                          & 28.45                          & 9.36 \\
NICE-SLAM \cite{Zhu2022CVPR}   & 28.80                          & 11.93                          & 36.03 \\
SDF-2-SDF \cite{slavcheva2018sdf} & 25.71                          & 10.05                          & 35.99 \\
\rowcolor[rgb]{ .847,  .847,  .847} Ours                           & 83.63                          & 67.52                          & 4.66 \bigstrut[b]\\

\Xhline{8\arrayrulewidth}
\end{tabular}%
}
\vspace{-0.1in}
\caption{Comparison on BEHAVE Dataset. ADD and ADD-S are AUC percentage (0 to 0.5~m).   Reconstruction is measured by chamfer distance.}
\label{tab:behave}
\vspace{-0.15in}
\end{table}

%% file: tables/ablations.tex
\begin{table}[h]
\vspace{-0.15in}
\centering
\def\mywidth{0.42\textwidth} 
\resizebox{\mywidth}{!}{
\begin{tabular}{l|rr|r}
\Xhline{8\arrayrulewidth}
\multicolumn{1}{c|}{\multirow{2}[2]{*}{Ablations}} & \multicolumn{2}{c|}{Pose} & \multicolumn{1}{l}{Reconstruction} \bigstrut[t]\\
      & ADD-S (\%) $\uparrow$ & ADD (\%) $\uparrow$ & CD (cm) $\downarrow$ \bigstrut[b]\\
\hline
Ours w/o memory & 82.05 & 56.96 & - \bigstrut[t]\\
Ours w/o NOF & 93.09 & 76.69 & - \\
Ours-GPG & 93.82 & 78.82 & - \\
Ours w/o hybrid SDF & 85.31 & 73.57 & 2.62 \\
Ours w/o compact mem pool & 87.48 & 59.99 & 0.90 \\
\rowcolor[rgb]{ .851,  .851,  .851} Ours  & \textbf{96.52} & \textbf{92.62} & \textbf{0.61} \bigstrut[b]\\
\Xhline{8\arrayrulewidth}
\end{tabular}%
}
\vspace{-0.1in}
\caption{Ablation study of our design choices. \textit{Ours w/o memory} removes the memory related modules and only performs frame-to-frame coarse pose estimation. \textit{Ours w/o NOF} removes the Neural Object Field module and $\mathcal{L}_s$ in Eq.~\eqref{eq:L_pg}.  \textit{Ours-GPG} replaces the Neural Object Field by global pose graph optimization using all memory frames. It runs in a separate thread concurrently  same as Neural Object Field. \textit{Ours w/o hybrid SDF} only considers foreground rays in the mask instead of hybrid SDF modeling. \textit{Ours w/o compact mem pool} adopts similar strategy of selecting frames to add into the memory pool as well as selecting the subset memory frames for pose graph optimization as in \cite{bundle2021wen}.}
\label{tab:ablations}
\vspace{-0.15in}
\end{table}

%% file: supplemental.tex
\section{Implementation Details}

During coarse pose initialization, if there is no immediate previous frame to compare with (\eg, missing detection by the segmentation, or object reappearing after complete occlusion), the current frame will instead be compared with the memory frames. The memory frame which has more than 10  feature correspondences with the current frame is selected as the new reference frame for the coarse pose initialization. The following steps remain the same.

For online pose graph optimization, we constrain the maximum number of participating memory frames $K=10$ for efficiency. When computing $\mathcal{L}_p$ we reject corresponding points whose distance is larger than 1 cm, or their normal angle is larger than 20\degree. The Gauss-Newton optimization iterates for 7 steps.

For Neural Object Field learning, we normalize the object into the neural volume bound of $[-1,1]$, where the scale is computed as 1.5 times of the initial frame's point cloud dimension. The neural volume's coordinate system is based on the  first frame's centered point cloud. The geometry network $\Omega$ consists of two-layer MLP with hidden dimension 64 and ReLU activation except for the last layer. The intermediate geometric feature $f_{\Omega(\cdot)}$ has dimension 16. The bias of the last layer is initialized to 0.1 for a small positive SDF prediction at the start of training. The appearance network $\Phi$ consists of three-layer MLP with hidden dimension 64 and ReLU activation except for the last layer, where we apply sigmoid activation to map the color prediction to $[0,1]$. For Octree ray-tracing, the finest voxel size is set to 2 cm. We simplify the multi-resolution hash encoder~\cite{mueller2022instant} to 4 levels, with number of feature vectors from 16 to 128 for efficiency. Each level's feature dimension is set to 2. The hash table size is set to $2^{22}$. In each iteration the ray batch size is 2048. For hierarchical point sampling, $N$ and $N'$ are set to 128 and 64, respectively.  The truncation distance $\lambda$ is set to 1 cm. For \textit{uncertain free space}, $\epsilon$ is set to 0.001. In the training loss, $w_u=100, w_e=1, w_{\textit{surf}}=1000, w_c=100, w_{\textit{eik}}=0.1$. We implement in PyTorch~\cite{paszke2019pytorch} with Adam optimizer. The initial learning rate is 0.01 with linear decay rate 0.1. The Neural Object Field training runs in a separate thread concurrently and interchanges data with the memory pool periodically after each training convergence (300 steps), which leads to sufficient pose refinement. The first training period starts when there are 10 memory frames in the pool. Upon training convergence, it returns the data to the memory pool and grabs memory frames newly added to the pool during its last training period, to repeat the training process. The next training reuses the latest updated frames' poses. But for the other trainable parameters, reusing their weights tend to get stuck in local minima if there is any sub-optimum in the previous training period, particularly due to noisy pose. Therefore, we re-initialize the network weights for the new training periods. This takes similar number of steps to refine the newly added memory frames' poses, compared to reusing the previous network weights.

\section{Computation Time}

All experiments were conducted on a standard
desktop with Intel i9-10980XE CPU
and a single NVIDIA RTX 3090 GPU. Our method consists of two threads running concurrently. The online tracking thread processes frames at around 10.2 Hz, where video segmentation takes 18 ms, coarse matching takes 24 ms, pose graph takes 56 ms
on average. Concurrently, the neural object field
thread runs in the background and takes 6.7 s averagely for each training round, at the end of which it exchanges data with the main thread. On the same hardware, competitive methods DROID-SLAM~\cite{teed2021droid} and BundleTrack~\cite{bundle2021wen} run at 6.1 Hz and 11.2 Hz respectively. 

\section{Metrics}

For evaluation, we decouple the pose estimation and shape reconstruction, so that they can be treated separately. For 6-DoF object pose evaluation, we compute the area under the curve (AUC) percentage of \textit{ADD} and \textit{ADD-S} metric: 
 \begin{equation}
     \text{ADD}=\frac{1}{|\mathcal{M}|}\sum_{x\in \mathcal{M}}^{}\left\| (Rx+t)-(\tilde{R}x+\tilde{t})  \right\|_2
 \end{equation}
  \begin{equation}
     \text{ADD-S}=\frac{1}{|\mathcal{M}|}\sum_{x_1\in \mathcal{M}} \minB_{x_2\in \mathcal{M}}\left\| (Rx_1+t)-(\tilde{R}x_2+\tilde{t})  \right\|_2,
 \end{equation}
 where $\mathcal{M}$ is the object model.
Since the novel unknown object's CAD model is inaccessible to the methods to define the coordinate system, we use the ground-truth pose in the first image to define the canonical coordinate frame of each video to evaluate the pose.

For 3D shape reconstruction evaluation, we report the results of chamfer distance between the final reconstructed mesh and the ground-truth mesh, using the following symmetric formulation:
\begin{align}
d_{CD}=&\frac{1}{2|\mathcal{M}_1|}\sum\limits_{x_1\in \mathcal{M}_1}^{}\min\limits_{x_2\in \mathcal{M}_2}\left\| x_1-x_2\right\|_2 + \\
&\frac{1}{2|\mathcal{M}_2|}\sum\limits_{x_2\in \mathcal{M}_2}^{}\min\limits_{x_1\in \mathcal{M}_1}\left\| x_1-x_2\right\|_2
\end{align}
In our method, the mesh can be extracted by applying Marching Cubes over the zero level set in the Neural Object Field. For all methods, we use the same resolution (5~mm) to sample points for evaluation. Since most videos do not cover the complete surrounding view of the object, we cull the ground-truth mesh faces that are never visible in the video by a rendering test, given by the ground-truth mesh and pose. 

\section{Detailed Results}

\begin{figure*}[hbt]
    \centering
    {\includegraphics[width=0.95\textwidth]{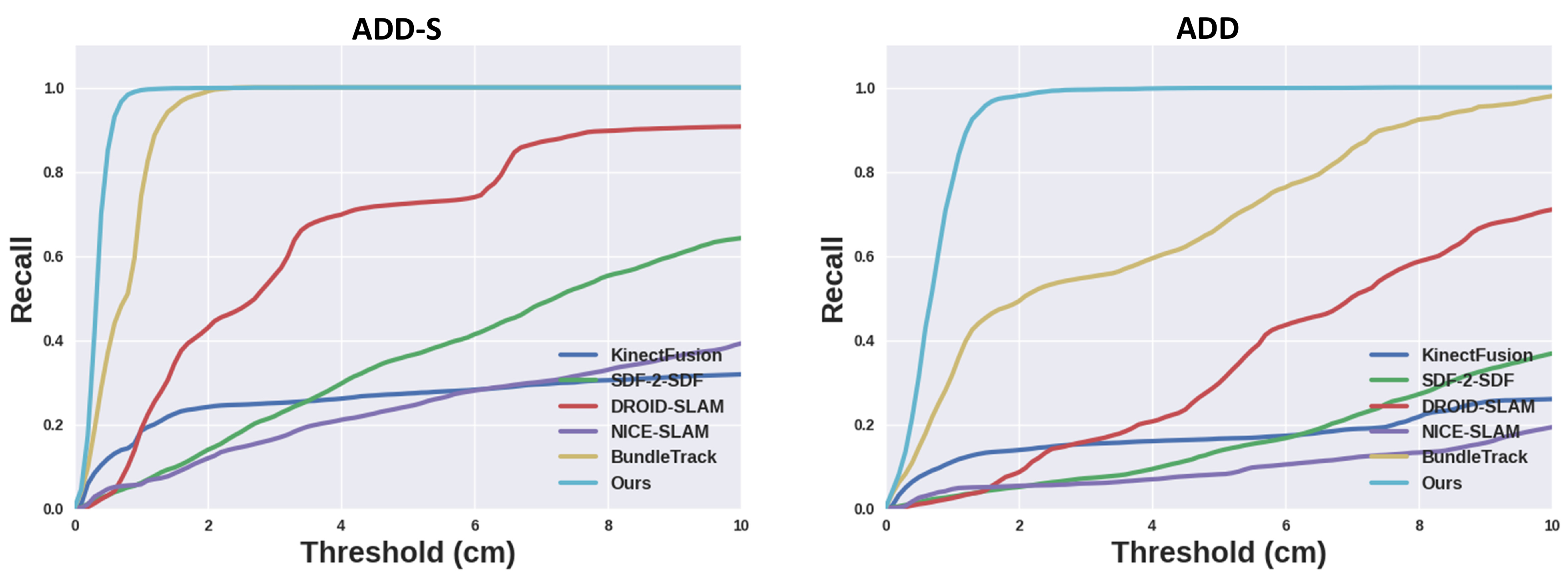}}
    \vspace{-0.1in}
    \caption{Recall curve of ADD-S (left) and ADD (right) metric  including all videos on HO3D Dataset.} \label{fig:ho3d_auc} 
\end{figure*}

\begin{figure*}[h]
    \centering
    {\includegraphics[width=0.95\textwidth]{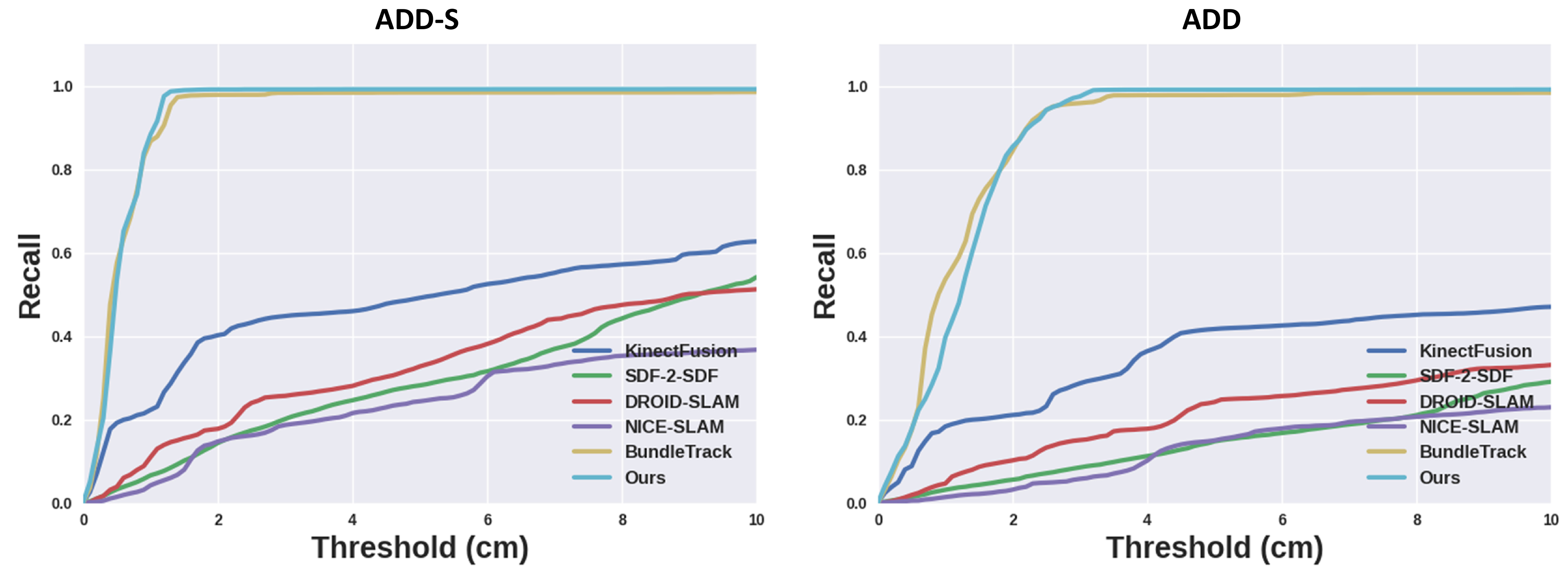}}
    \vspace{-0.1in}
    \caption{Recall curve of ADD-S (left) and ADD (right) metric  including all videos on YCBInEOAT Dataset.}  \label{fig:ycb_auc}
\end{figure*}

\begin{figure*}[h]
    \centering
    {\includegraphics[width=0.95\textwidth]{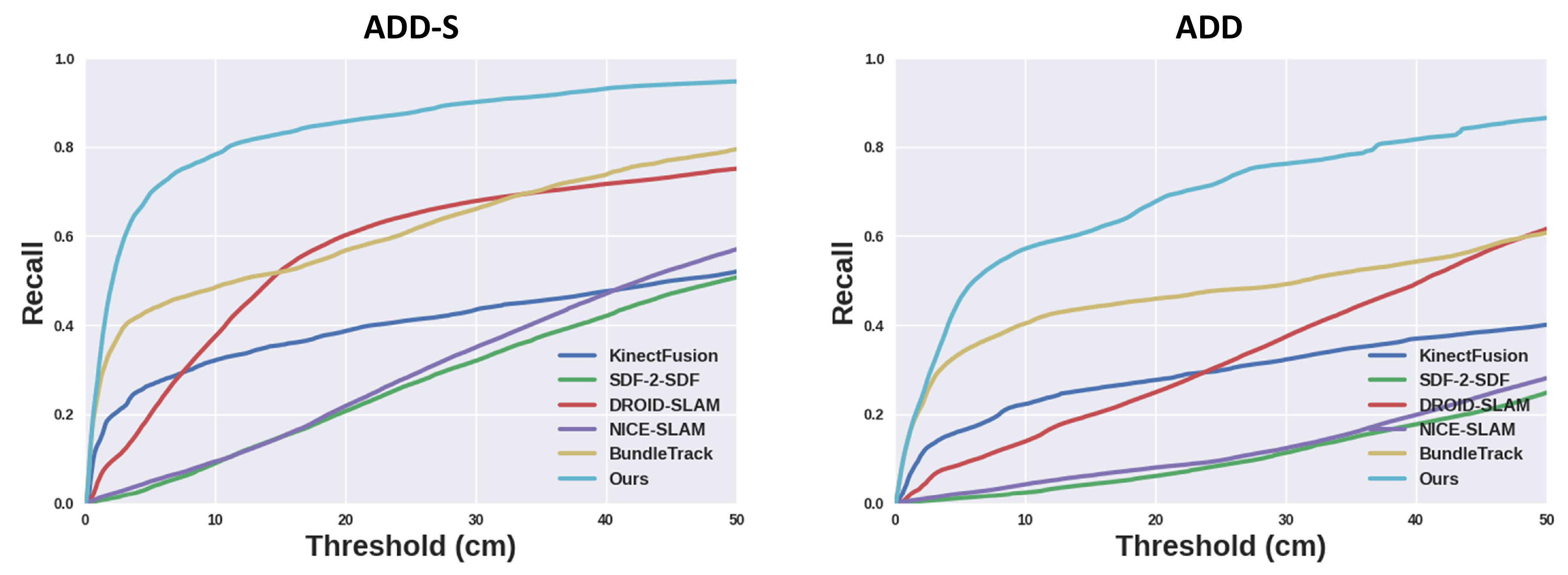}}
    \vspace{-0.1in}
    \caption{Recall curve of ADD-S (left) and ADD (right) metric  including all videos on BEHAVE Dataset.} \label{fig:behave_auc} 
\end{figure*}

\textbf{Recall curves} for ADD-S and ADD for all three datasets are presented in Fig.~\ref{fig:ho3d_auc} (HO3D), Fig.~\ref{fig:ycb_auc} (YCBInEOAT), and Fig.~\ref{fig:behave_auc} (BEHAVE). 
Each plot shows the results for all videos of the respective dataset.
As can be seen, the area-under-the-curve (AUC) for our method exceeds that of other methods for almost all datasets.

\textbf{Per-video quantitative results} for all three datasets are presented in Tab.~\ref{tab:ho3d_detail} (HO3D), Tab.~\ref{tab:ycbineoat_detail} (YCBInEOAT), and Tabs.~\ref{tab:behave_detailA}-\ref{tab:behave_detailD} (BEHAVE).  
As can be seen, our method performs best on almost all videos of HO3D, more than half the videos of YCBInEOAT, and a large majority of videos of BEHAVE.
Note that the last row of each table (``Mean'') is included in the main paper.

\begin{figure*}[h]
    \centering
    {\includegraphics[width=0.9\textwidth]{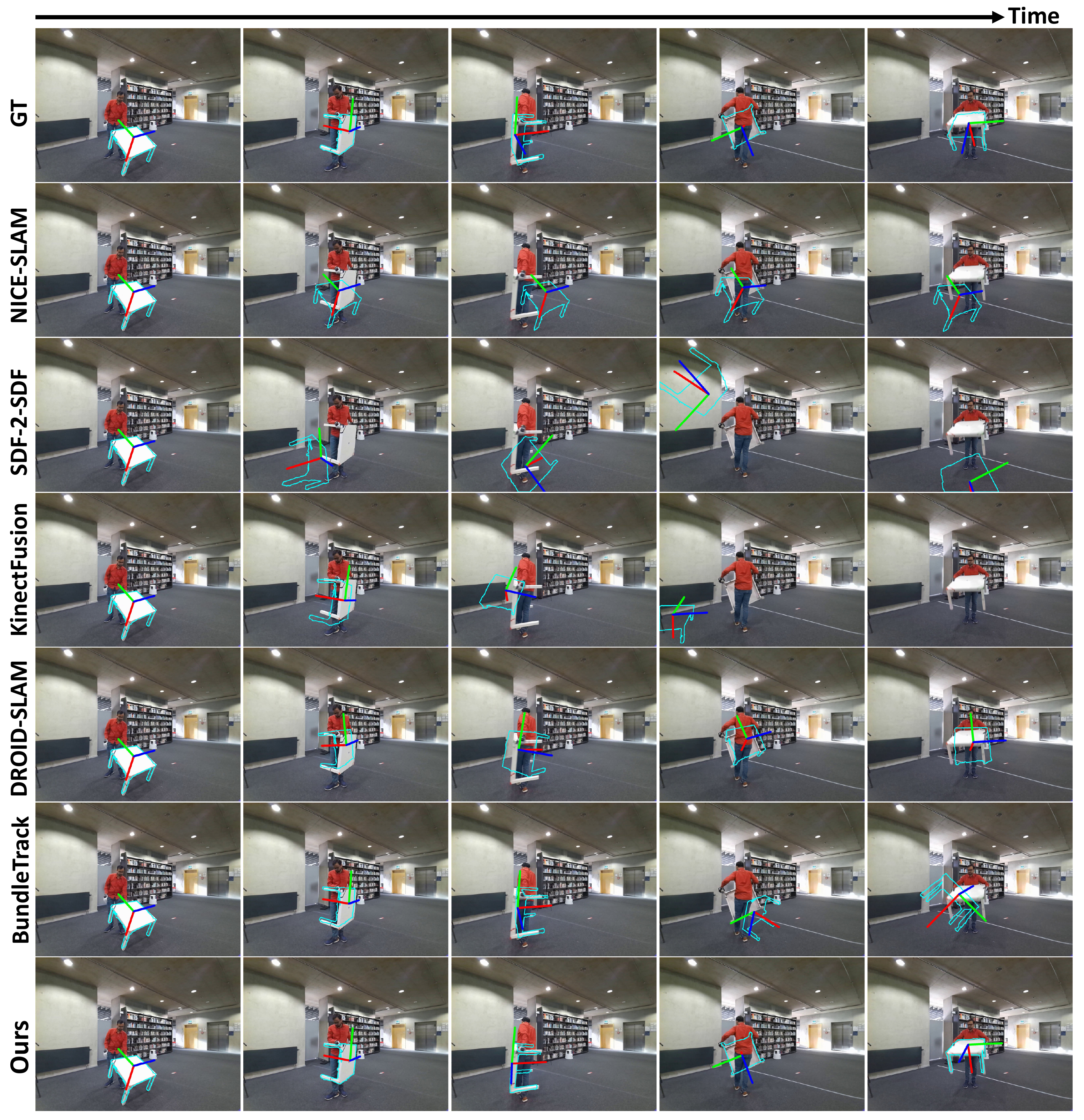}}
    \vspace{-0.1in}
    \caption{Qualitative comparison on BEHAVE video ``Date03\_Sub04\_tablesquare\_lift.3''. Our method is sometimes even more accurate than ground-truth (3rd and last column). It is also robust to severe occlusions (4th column).}\label{fig:behave_table_comparison}
\end{figure*}

\textbf{Qualitative results} are demonstrated in Figs.~\ref{fig:ho3d_AP_13_comparison} and~\ref{fig:ho3d_MPM13_comparison} (HO3D), 
Fig.~\ref{fig:ycb_sugar_box_comparison} (YCBInEOAT),
and
Figs.~\ref{fig:behave_chair_comparison} and \ref{fig:behave_table_comparison} (BEHAVE). 
We encourage the reader to watch the supplemental video.

\vspace{8pt}\noindent \textbf{Details Regarding the Single-View Setup of BEHAVE.} As mentioned in the paper, the BEHAVE Dataset was captured by a pre-calibrated multi-camera system with four cameras. 
Since our method only requires a monocular input, for fair evaluation, we run all methods on a single monocular input.  
That is, for each scene, we input only one of the cameras' captured video to the methods. 

Although in theory we could run each method four times, once per video camera, this would be excessively time consuming for the little insight that it might bring.
Moreover, since there are only four cameras placed at each corner around the scene, it is often the case that the object is severely occluded by the human in several cameras' views (including at the beginning of the video).  Using such cameras would not lead to meaningful results for tracking evaluation, due to the very limited object visibility at initialization. 

Instead, we decided to automatically select one of the four cameras from each scene for evaluation.
More specifically, we select the video with the least amount of occlusion in each scene over the entire sequence. To do so, we compute the average visibility ratio of the object in each camera's video by comparing the ground-truth object mask against the rendered object mask using the ground-truth information. This is performed offline for all videos before evaluation.  The selected single-view video is then used by all methods for evaluation, even though severe occlusions still occur frequently which exhibit challenges, as shown in Fig.~\ref{fig:behave_chair_comparison}, \ref{fig:behave_table_comparison}.

\section{Robustness Analysis}

In the following we discuss our approach's robustness under various challenges.  We encourage the reader to watch our supplemental video for more complete appreciation of the system.

\begin{figure*}[h]
    \centering
    {\includegraphics[width=0.9\textwidth]{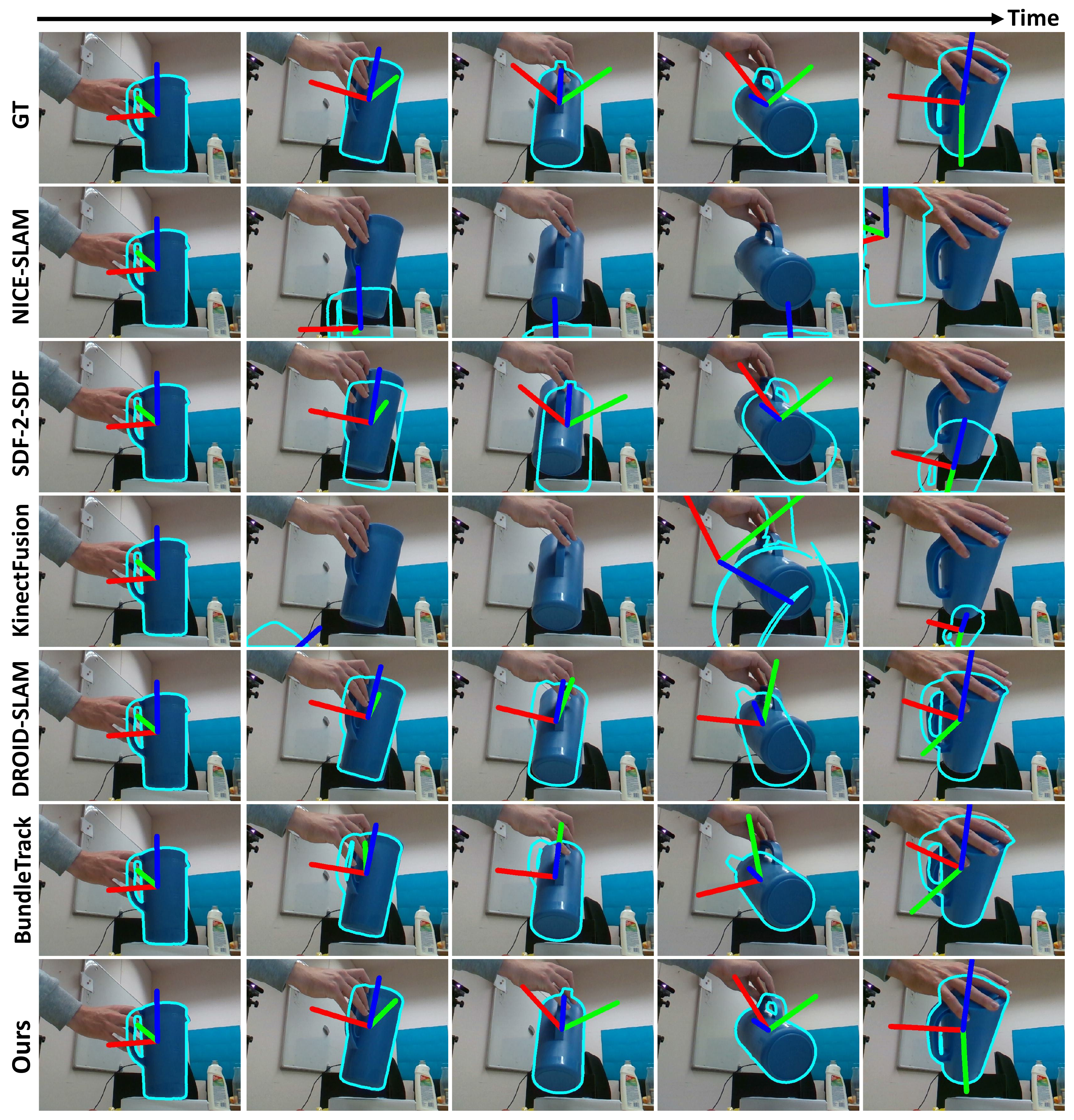}}
    \vspace{-0.1in}
    \caption{Qualitative comparison on HO3D video ``AP13''. Our method is robust to observations with little texture or geometric cues (large area of cylindrical surface), whereas comparison methods struggle.} \label{fig:ho3d_AP_13_comparison}
\end{figure*}

\begin{figure*}[h]
    \centering
    {\includegraphics[width=0.9\textwidth]{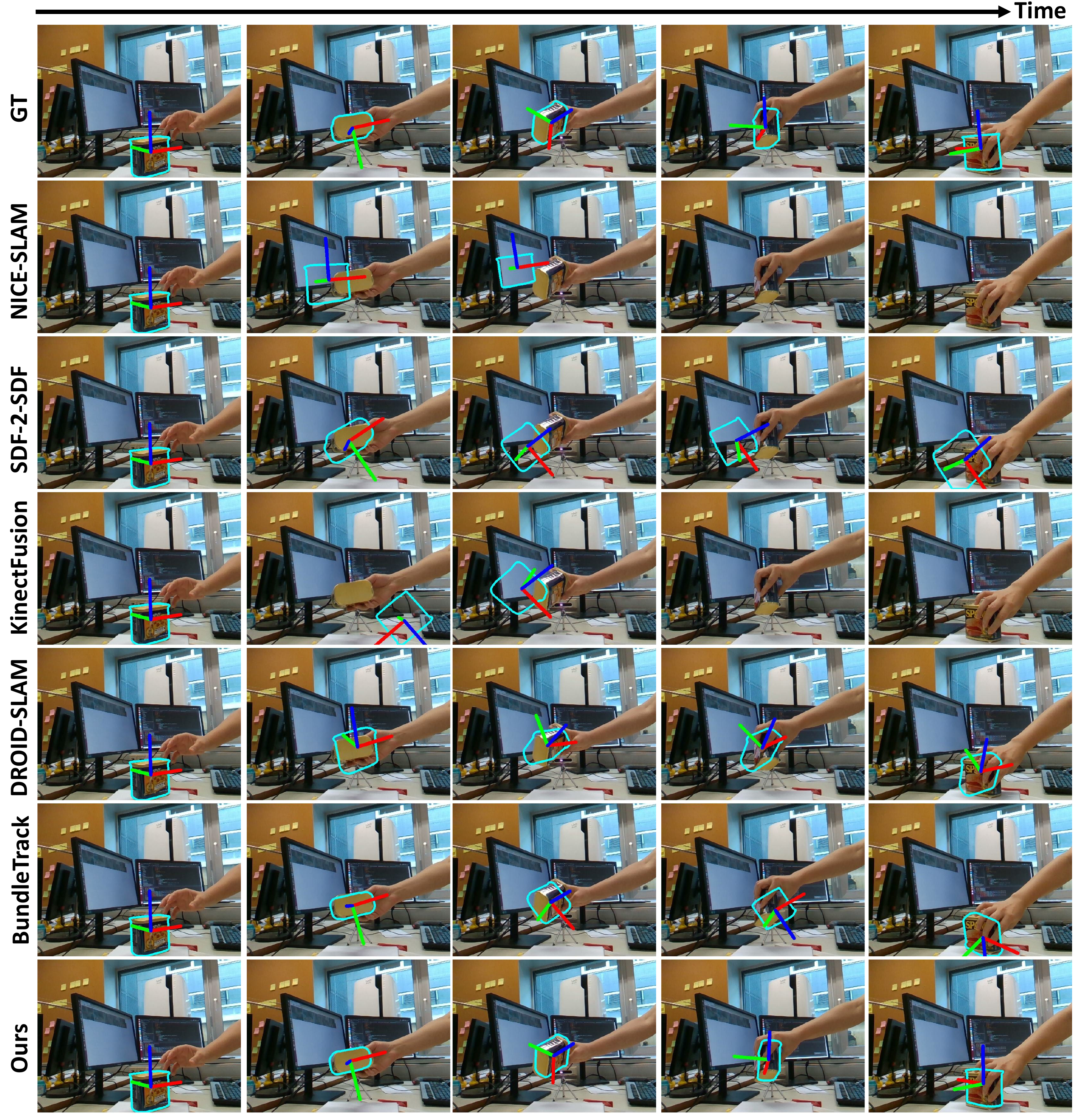}}
    \vspace{-0.1in}
    \caption{Qualitative comparison on HO3D video ``MPM13''. Note that our pose tracking at times appears to be slightly more accurate than the ground-truth as shown in the rightmost column.}\label{fig:ho3d_MPM13_comparison}
\end{figure*}

\begin{figure*}[h]
    \centering
    {\includegraphics[width=0.9\textwidth]{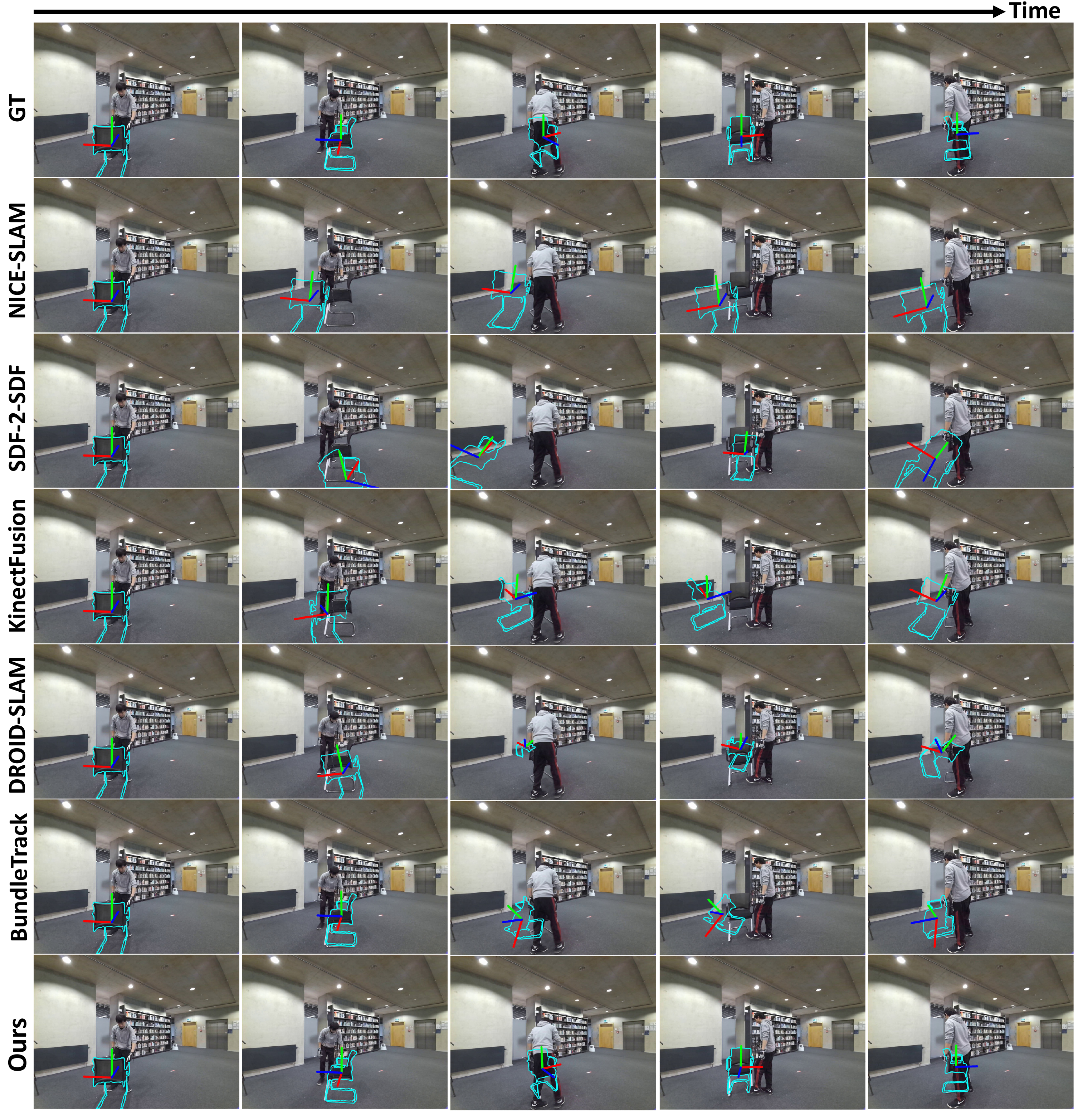}}
    \vspace{-0.1in}
    \caption{Qualitative comparison on BEHAVE video ``Date03\_Sub03\_chairblack\_hand.3''. Our method is robust to severe and even complete occlusions (3rd and last column).}\label{fig:behave_chair_comparison}
\end{figure*}

\vspace{6pt}\noindent \textbf{Dearth of Texture or Geometric Cues.} In the case of dynamic object-centric setting, dearth of texture or geometric cues frequently occur given by the object itself. For instance, in Fig.~\ref{fig:ho3d_AP_13_comparison}, large areas on the blue pitcher lack texture, which challenge those methods heavily relying on optical flow (DROID-SLAM~\cite{teed2021droid}), or keypoint matching (BundleTrack~\cite{bundle2021wen}), or photometric loss (NICE-SLAM~\cite{Zhu2022CVPR}). Additionally, large areas of cylindrical surface also exhibit few geometric cues to leverage and can cause rotational ambiguity to those methods relying on point-to-surface matching (SDF2SDF~\cite{slavcheva2018sdf}, BundleTrack~\cite{bundle2021wen}, KinectFusion~\cite{newcombe2011kinectfusion}). In contrast, our method is robust to these challenges due to the synergy of  pose graph optimization and Neural Object Field. More examples of such challenges can be found in Fig.~\ref{fig:ho3d_MPM13_comparison}, \ref{fig:behave_chair_comparison},  \ref{fig:behave_table_comparison}.

\begin{figure*}[h]
    \centering
    {\includegraphics[width=0.9\textwidth]{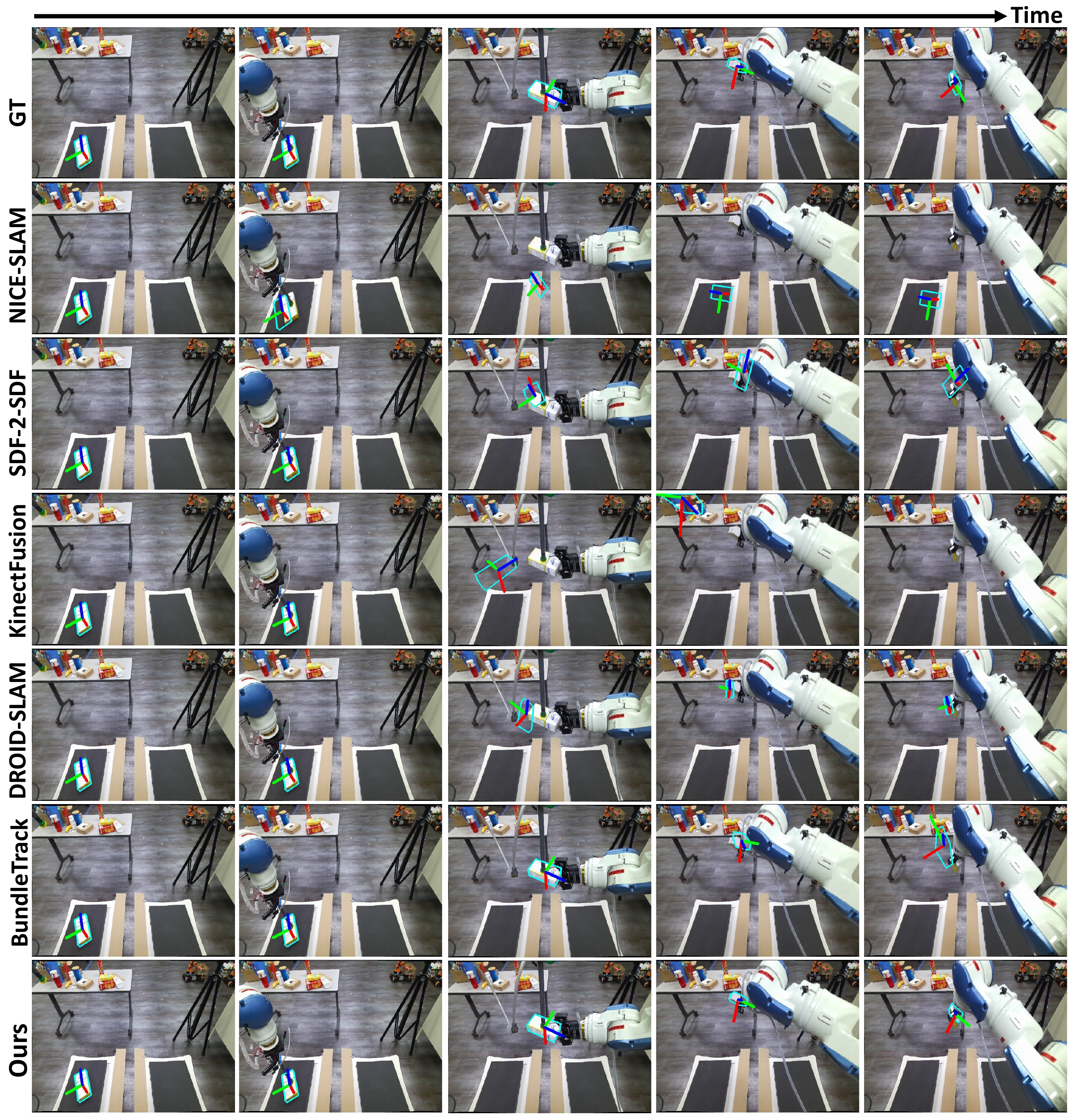}}
    \vspace{-0.1in}
    \caption{Qualitative comparison on YCBInEOAT video ``sugar\_box1".}\label{fig:ycb_sugar_box_comparison}
\end{figure*}

\vspace{6pt}\noindent \textbf{Occlusions.} In the dynamic object setting, occlusions  include self-occlusions and external occlusions introduced by the interaction agent (\eg, human hand, human body, robotic arm). For instance, in Fig.~\ref{fig:ho3d_MPM13_comparison}, there are moments when the ``meat can'' only exhibits a single flat face (2nd column) after extreme rotations, causing severe self-occlusion. In other observations, external occlusion introduced by the human hand (4th column) also challenges the comparison methods. More examples of such challenges can be found in Fig.~\ref{fig:ho3d_AP_13_comparison}, \ref{fig:behave_chair_comparison}, \ref{fig:behave_table_comparison},
\ref{fig:ycb_sugar_box_comparison}. As can be observed, our method is robust to either case and keeps tracking accurately throughout the video thanks to the memory mechanism, whereas the comparison methods struggle. 

\vspace{6pt}\noindent \textbf{Specularity.} Due to the object's surface smoothness, material and complex environmental lighting, specularity could happen, introducing challenges for those methods heavily replying on optical flow (DROID-SLAM~\cite{teed2021droid}), keypoint matching (BundleTrack~\cite{bundle2021wen}) or photometric loss (NICE-SLAM~\cite{Zhu2022CVPR}). As shown in Fig.~\ref{fig:ho3d_AP_13_comparison}, \ref{fig:ho3d_MPM13_comparison}, \ref{fig:behave_chair_comparison}, \ref{fig:ycb_sugar_box_comparison}, despite the specularity on metalic or highly smooth surfaces, our method keeps tracking accurately throughout the video, whereas the comparison methods become brittle. 

\begin{figure*}[h]
    \centering
    {\includegraphics[width=0.95\textwidth]{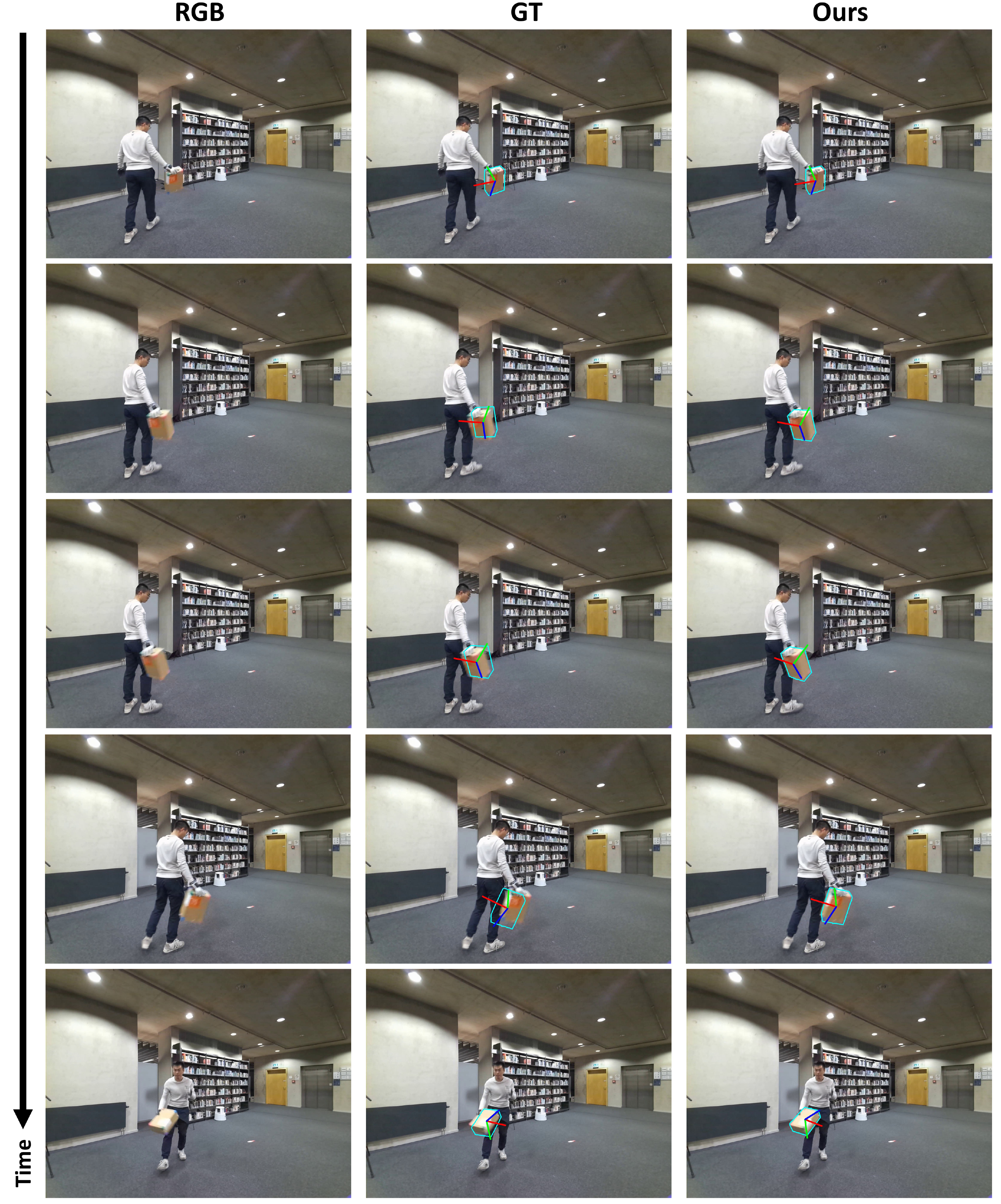}}
    \vspace{-0.1in}
    \caption{Despite fast object pose change and motion blur, our approach  produces even more accurate pose than ground-truth. Image is best viewed by zooming in.} \label{fig:behave_motion_blur} 
\end{figure*}

\vspace{6pt}\noindent \textbf{Abrupt Motion and Motion Blur.} Fig.~\ref{fig:behave_motion_blur} illustrates an example of abrupt object motion due to the human freely swinging the box. Aside from challenges for 6-DoF pose tracking under large displacement, it causes motion blur in RGB, leading to additional challenge for keypoint matching and Neural Object Field learning. However, our method has shown robustness under these adverse conditions and even yields more accurate pose than ground-truth.

\begin{figure*}[h]
    \centering
    {\includegraphics[width=0.95\textwidth]{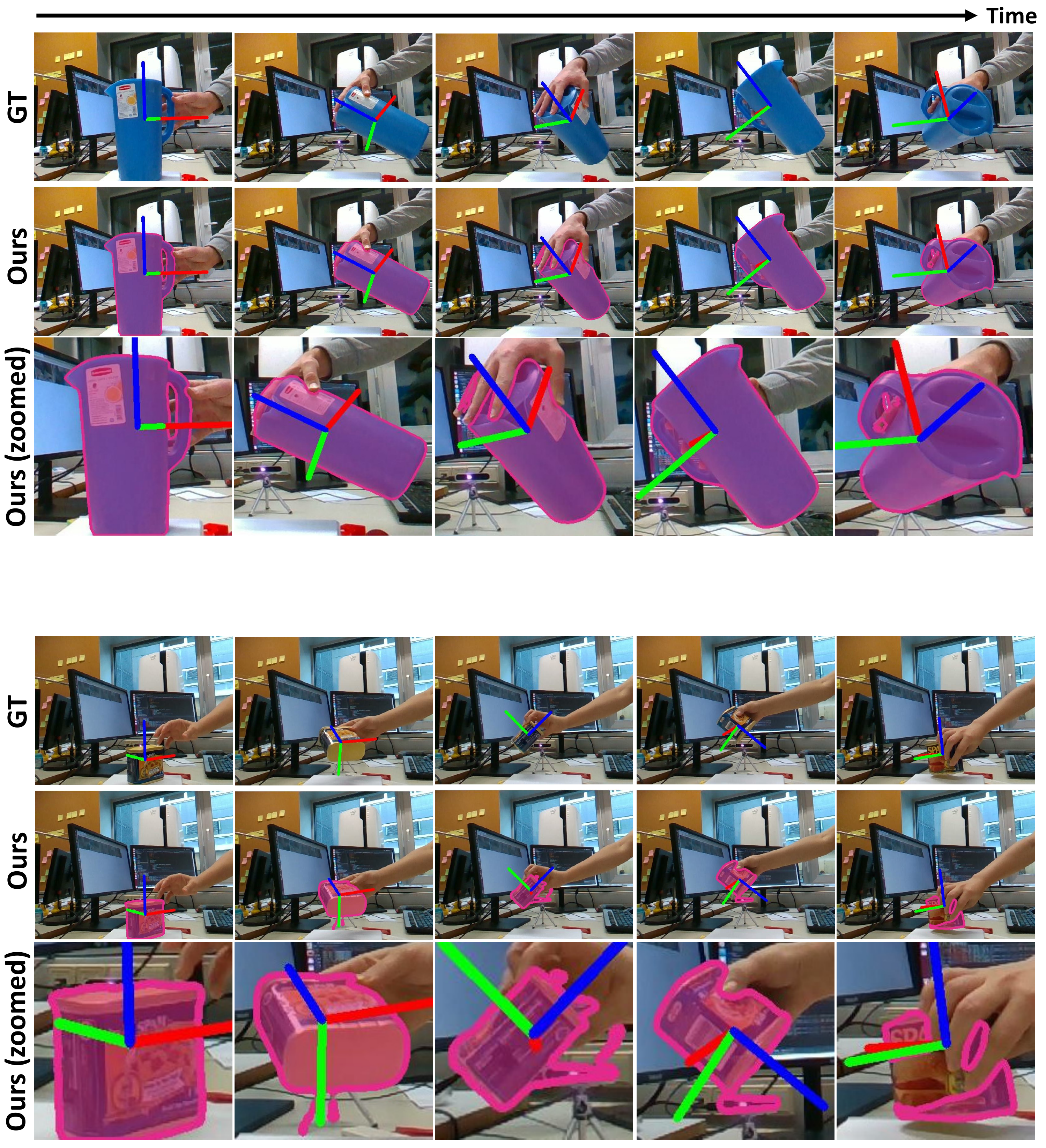}}
    \vspace{-0.1in}
    \caption{Example of noisy masks (purple) from the video segmentation network, showing both false positive and false negative predictions.  The first column visualizes the first frame's mask that initializes tracking.  Our method is robust to noisy segmentation and maintains accurate tracking despite such noise.  Figure is continued on the next page.  (Part 1 of 2.)} \label{fig:noisy_segA} 
\end{figure*}

\begin{figure*}[h]
    \centering
    {\includegraphics[width=0.95\textwidth]{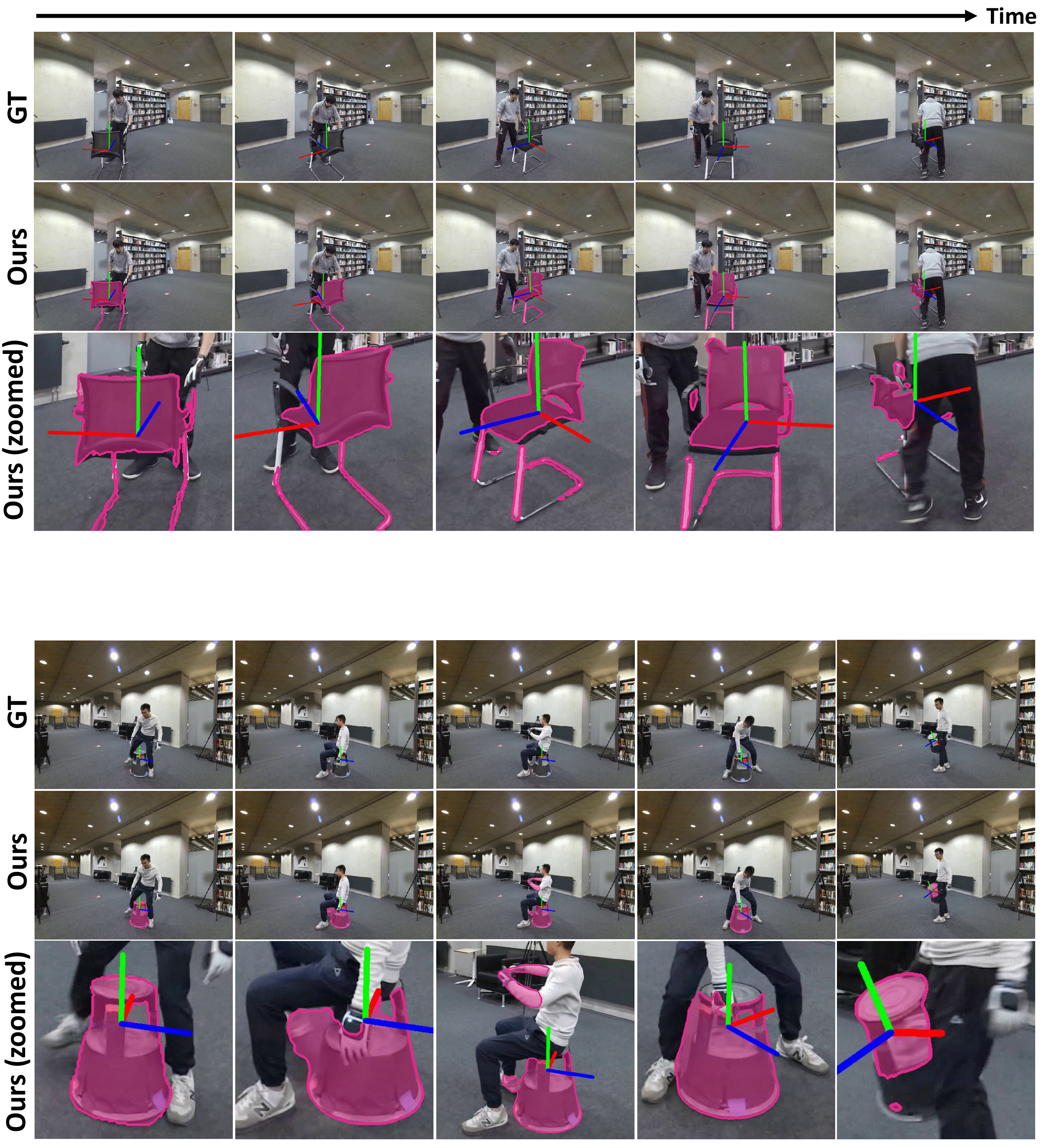}}
    \vspace{-0.1in}
    \caption{Example of noisy masks (purple) from the video segmentation network.  Continued from previous figure.  (Part 2 of 2.)} \label{fig:noisy_segB} 
\end{figure*}

\vspace{6pt}\noindent \textbf{Noisy Segmentation.} Figs.~\ref{fig:noisy_segA} and \ref{fig:noisy_segB} demonstrate examples of noisy  masks (purple) from the video segmentation network, including both false positive and false negative predictions. The false negative segmentation leads to ignorance of the texture-rich areas, intensifying the issue of dearth of texture. The false positive segmentation introduces deformable part from the interaction agent or undesired scene background, causing inconsistency in multi-view. However, our downstream modules are robust to the segmentation noise and maintain accurate tracking.

\begin{figure*}[h]
    \centering
    {\includegraphics[width=0.95\textwidth]{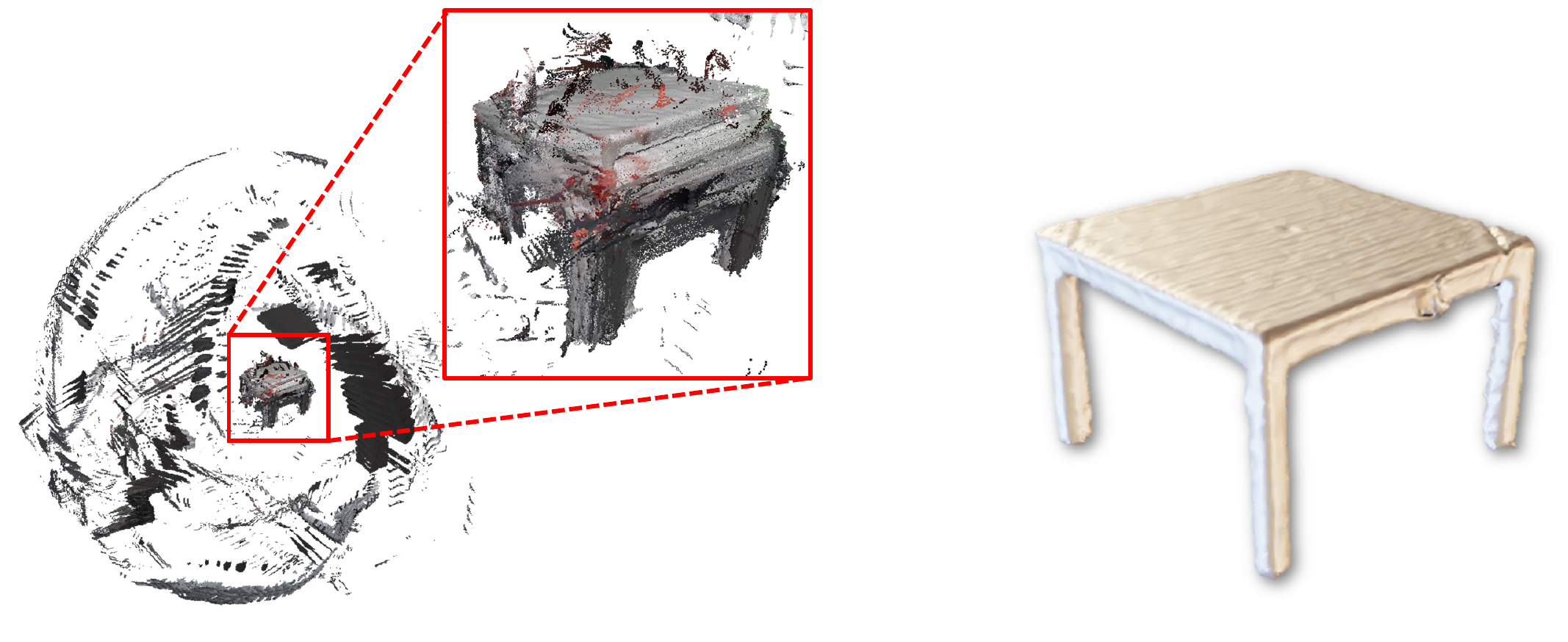}}
    \vspace{-0.1in}
    \caption{Example of noisy depth from BEHAVE video ``Date03\_Sub04\_tablesquare\_lift.3''. \textbf{Left:} Fused point cloud using ground-truth pose and masks from the video segmentation network. \textbf{Right:} Final reconstruction from our approach without any trimming.} \label{fig:noisy_depth} 
\end{figure*}

\vspace{6pt}\noindent \textbf{Noisy Depth.} As shown in Fig.~\ref{fig:noisy_depth}, in our setting, the noisy depth comes from two sources. First, the consumer-level RGBD camera has observable sensing noise. This is especially the case for BEHAVE~\cite{bhatnagar22behave} and YCBInEOAT~\cite{wen2020se} Dataset, where the images are captured at a distance from the camera, which challenges depth sensing. Second, due to the noisy segmentation, false positive predictions include undesired background areas in the depth point cloud. In Fig.~\ref{fig:noisy_depth} (left), when naively fusing the per-frame depth point cloud using ground-truth pose, the result is highly cluttered, which implies the noisy depth sensing and segmentation. However, despite such noise, our simultaneous pose tracking and reconstruction produce high quality mesh, as shown on the right.

\section{Limitation and Failure Modes}

\begin{figure*}[h]
    \centering
    {\includegraphics[width=0.95\textwidth]{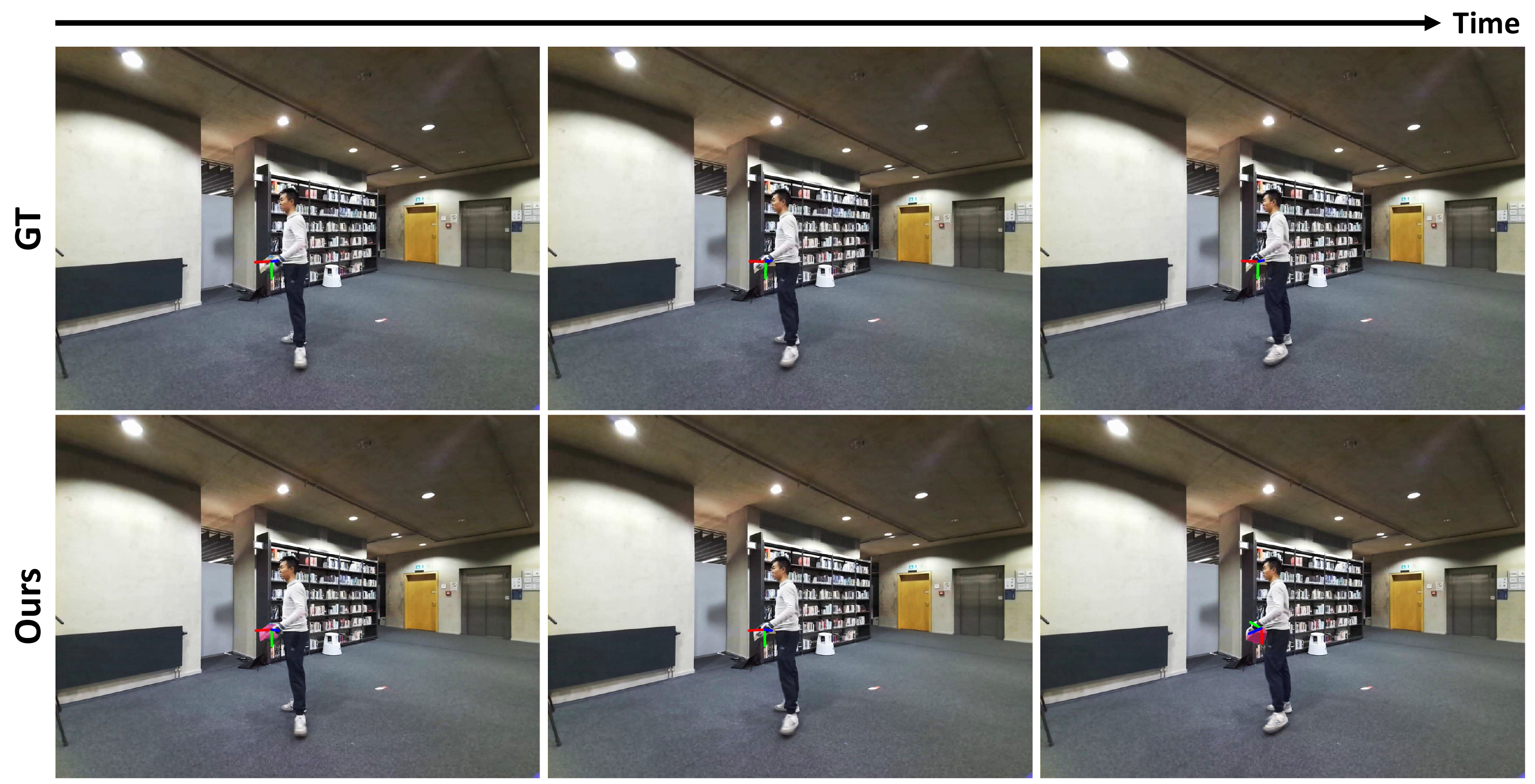}}
    \vspace{-0.1in}
    \caption{Failure case. The occurrence of severe occlusion, segmentation error, dearth of texture or geometric cues together lead to tracking failure. When the object re-appears, the recovered pose is affected by symmetric geometry.} \label{fig:failure} 
\end{figure*}

While our method is robust to a variety of challenging conditions, it fails when multiple types of challenges appear together. For instance, in Fig.~\ref{fig:failure}, the occurrence of severe occlusion, segmentation error, dearth of texture and geometric cues together lead to tracking failure. When the object re-appears, the recovered pose is affected by symmetric geometry. Besides, our method requires depth modality which limits its application to certain types of objects where depth sensing fails, such as transparent objects. Finally, our method assumes the object to be rigid. In future work, generalizing to both rigid and non-rigid objects at the same time would be of interest.

\input{tables/ho3d_detailed_quant.tex}

\input{tables/ycbineoat_detailed_quant.tex}

\input{tables/behave_detailed_quantA.tex}
\input{tables/behave_detailed_quantB.tex}
\input{tables/behave_detailed_quantC.tex}
\input{tables/behave_detailed_quantD.tex}

\clearpage

\clearpage

\clearpage

\clearpage

\clearpage

%% file: tables/ho3d_detailed_quant.tex
\clearpage

\begin{table*}[h]
\centering
\def\mywidth{0.9\textwidth} 
\resizebox{\mywidth}{!}{
\begin{tabular}{c|c|cccccc}
\tableborder
Video                          & Metric                         & DROID-SLAM \cite{teed2021droid} & BundleTrack \cite{bundle2021wen} & KinectFusion \cite{newcombe2011kinectfusion} & NICE-SLAM \cite{Zhu2022CVPR}   & SDF-2-SDF \cite{slavcheva2018sdf} & \cellcolor[rgb]{ .851,  .851,  .851}Ours \bigstrut[b]\\
\hline
\multicolumn{1}{c|}{\multirow{3}[2]{*}{AP10}} & ADD-S (\%) $\uparrow$          & 89.36                          & 91.68                          & 11.39                          & 14.11                          & 33.54                          & \cellcolor[rgb]{ .851,  .851,  .851}\textbf{96.10} \bigstrut[t]\\
                               & ADD (\%) $\uparrow$            & 50.06                          & 36.60                          & 9.99                           & 2.62                           & 16.35                          & \cellcolor[rgb]{ .851,  .851,  .851}\textbf{91.00} \\
                               & CD (cm) $\downarrow$           & 2.48                           & 1.88                           & 4.18                           & 33.22                          & 12.01                          & \cellcolor[rgb]{ .851,  .851,  .851}\textbf{0.47} \bigstrut[b]\\
\hline
\multicolumn{1}{c|}{\multirow{3}[2]{*}{AP11}} & ADD-S (\%) $\uparrow$          & 68.76                          & 91.45                          & 76.34                          & 11.40                          & 21.21                          & \cellcolor[rgb]{ .851,  .851,  .851}\textbf{96.18} \bigstrut[t]\\
                               & ADD (\%) $\uparrow$            & 26.24                          & 41.28                          & 30.99                          & 3.62                           & 7.65                           & \cellcolor[rgb]{ .851,  .851,  .851}\textbf{91.76} \\
                               & CD (cm) $\downarrow$           & 120.91                         & 129.18                         & 21.65                          & 90.13                          & 16.79                          & \cellcolor[rgb]{ .851,  .851,  .851}\textbf{0.56} \bigstrut[b]\\
\hline
\multicolumn{1}{c|}{\multirow{3}[2]{*}{AP12}} & ADD-S (\%) $\uparrow$          & 38.71                          & 90.79                          & 20.52                          & 19.90                          & 19.48                          & \cellcolor[rgb]{ .851,  .851,  .851}\textbf{97.06} \bigstrut[t]\\
                               & ADD (\%) $\uparrow$            & 7.15                           & 50.82                          & 9.13                           & 4.11                           & 2.78                           & \cellcolor[rgb]{ .851,  .851,  .851}\textbf{94.76} \\
                               & CD (cm) $\downarrow$           & 10.43                          & 2.47                           & 17.18                          & 52.11                          & 8.66                           & \cellcolor[rgb]{ .851,  .851,  .851}\textbf{0.59} \bigstrut[b]\\
\hline
\multicolumn{1}{c|}{\multirow{3}[2]{*}{AP13}} & ADD-S (\%) $\uparrow$          & 91.68                          & 90.68                          & 11.40                          & 32.45                          & 49.90                          & \cellcolor[rgb]{ .851,  .851,  .851}\textbf{96.16} \bigstrut[t]\\
                               & ADD (\%) $\uparrow$            & 73.67                          & 49.03                          & 9.46                           & 6.11                           & 18.16                          & \cellcolor[rgb]{ .851,  .851,  .851}\textbf{92.73} \\
                               & CD (cm) $\downarrow$           & 3.00                           & 2.77                           & 19.76                          & 37.62                          & 12.22                          & \cellcolor[rgb]{ .851,  .851,  .851}\textbf{0.63} \bigstrut[b]\\
\hline
\multicolumn{1}{c|}{\multirow{3}[2]{*}{AP14}} & ADD-S (\%) $\uparrow$          & 35.53                          & \textbf{96.02}                 & 18.43                          & 5.98                           & 45.56                          & \cellcolor[rgb]{ .851,  .851,  .851}96.01 \bigstrut[t]\\
                               & ADD (\%) $\uparrow$            & 0.06                           & 90.30                          & 15.81                          & 0.34                           & 32.54                          & \cellcolor[rgb]{ .851,  .851,  .851}\textbf{91.25} \\
                               & CD (cm) $\downarrow$           & 71.68                          & 72.40                          & 20.92                          & 31.91                          & 4.38                           & \cellcolor[rgb]{ .851,  .851,  .851}\textbf{1.28} \bigstrut[b]\\
\hline
\multicolumn{1}{c|}{\multirow{3}[2]{*}{MPM10}} & ADD-S (\%) $\uparrow$          & 0.33                           & 94.94                          & 12.82                          & 29.20                          & 41.85                          & \cellcolor[rgb]{ .851,  .851,  .851}\textbf{95.05} \bigstrut[t]\\
                               & ADD (\%) $\uparrow$            & 0.27                           & 87.45                          & 9.37                           & 7.17                           & 15.23                          & \cellcolor[rgb]{ .851,  .851,  .851}\textbf{88.92} \\
                               & CD (cm) $\downarrow$           & 1.38                           & 0.97                           & 16.81                          & 54.71                          & 5.86                           & \cellcolor[rgb]{ .851,  .851,  .851}\textbf{0.56} \bigstrut[b]\\
\hline
\multicolumn{1}{c|}{\multirow{3}[2]{*}{MPM11}} & ADD-S (\%) $\uparrow$          & 59.68                          & 89.94                          & 13.10                          & 5.34                           & 13.06                          & \cellcolor[rgb]{ .851,  .851,  .851}\textbf{96.20} \bigstrut[t]\\
                               & ADD (\%) $\uparrow$            & 20.32                          & 53.20                          & 9.74                           & 3.55                           & 6.15                           & \cellcolor[rgb]{ .851,  .851,  .851}\textbf{91.51} \\
                               & CD (cm) $\downarrow$           & 87.41                          & 88.97                          & 15.72                          & 66.32                          & 6.82                           & \cellcolor[rgb]{ .851,  .851,  .851}\textbf{0.49} \bigstrut[b]\\
\hline
\multicolumn{1}{c|}{\multirow{3}[2]{*}{MPM12}} & ADD-S (\%) $\uparrow$          & 84.43                          & 95.66                          & 12.59                          & 3.99                           & 26.08                          & \cellcolor[rgb]{ .851,  .851,  .851}\textbf{96.98} \bigstrut[t]\\
                               & ADD (\%) $\uparrow$            & 53.29                          & 90.96                          & 6.70                           & 0.35                           & 8.48                           & \cellcolor[rgb]{ .851,  .851,  .851}\textbf{93.13} \\
                               & CD (cm) $\downarrow$           & 1.70                           & 121.33                         & 15.92                          & 51.38                          & 10.24                          & \cellcolor[rgb]{ .851,  .851,  .851}\textbf{0.46} \bigstrut[b]\\
\hline
\multicolumn{1}{c|}{\multirow{3}[2]{*}{MPM13}} & ADD-S (\%) $\uparrow$          & 75.30                          & 89.42                          & 10.58                          & 14.34                          & 40.95                          & \cellcolor[rgb]{ .851,  .851,  .851}\textbf{95.80} \bigstrut[t]\\
                               & ADD (\%) $\uparrow$            & 22.61                          & 38.78                          & 7.27                           & 6.67                           & 9.49                           & \cellcolor[rgb]{ .851,  .851,  .851}\textbf{90.62} \\
                               & CD (cm) $\downarrow$           & 3.27                           & 81.39                          & 18.41                          & 72.50                          & 6.05                           & \cellcolor[rgb]{ .851,  .851,  .851}\textbf{0.57} \bigstrut[b]\\
\hline
\multicolumn{1}{c|}{\multirow{3}[2]{*}{MPM14}} & ADD-S (\%) $\uparrow$          & 73.46                          & 95.49                          & 26.70                          & 76.36                          & 46.19                          & \cellcolor[rgb]{ .851,  .851,  .851}\textbf{97.33} \bigstrut[t]\\
                               & ADD (\%) $\uparrow$            & 26.12                          & 90.16                          & 11.05                          & 26.94                          & 20.57                          & \cellcolor[rgb]{ .851,  .851,  .851}\textbf{94.52} \\
                               & CD (cm) $\downarrow$           & 6.50                           & 94.99                          & 12.52                          & 52.84                          & 6.18                           & \cellcolor[rgb]{ .851,  .851,  .851}\textbf{0.47} \bigstrut[b]\\
\hline
\multicolumn{1}{c|}{\multirow{3}[2]{*}{SB11}} & ADD-S (\%) $\uparrow$          & 63.39                          & 94.44                          & 58.72                          & 30.06                          & 9.67                           & \cellcolor[rgb]{ .851,  .851,  .851}\textbf{97.27} \bigstrut[t]\\
                               & ADD (\%) $\uparrow$            & 32.15                          & 84.64                          & 55.25                          & 23.72                          & 5.93                           & \cellcolor[rgb]{ .851,  .851,  .851}\textbf{94.39} \\
                               & CD (cm) $\downarrow$           & 84.72                          & 75.83                          & 3.01                           & 81.73                          & 20.19                          & \cellcolor[rgb]{ .851,  .851,  .851}\textbf{0.46} \bigstrut[b]\\
\hline
\multicolumn{1}{c|}{\multirow{3}[2]{*}{SB13}} & ADD-S (\%) $\uparrow$          & 91.88                          & 95.66                          & 32.15                          & 36.05                          & 47.73                          & \cellcolor[rgb]{ .851,  .851,  .851}\textbf{97.67} \bigstrut[t]\\
                               & ADD (\%) $\uparrow$            & 76.44                          & 85.47                          & 30.89                          & 26.74                          & 32.50                          & \cellcolor[rgb]{ .851,  .851,  .851}\textbf{95.24} \\
                               & CD (cm) $\downarrow$           & 3.15                           & 2.49                           & 21.39                          & 32.91                          & 9.60                           & \cellcolor[rgb]{ .851,  .851,  .851}\textbf{0.47} \bigstrut[b]\\
\hline
\multicolumn{1}{c|}{\multirow{3}[2]{*}{SM1}} & ADD-S (\%) $\uparrow$          & 67.86                          & 84.94                          & 30.88                          & 10.65                          & 71.19                          & \cellcolor[rgb]{ .851,  .851,  .851}\textbf{96.90} \bigstrut[t]\\
                               & ADD (\%) $\uparrow$            & 45.25                          & 59.41                          & 9.41                           & 4.64                           & 33.19                          & \cellcolor[rgb]{ .851,  .851,  .851}\textbf{94.24} \\
                               & CD (cm) $\downarrow$           & 4.21                           & 2.04                           & 13.95                          & 26.05                          & 6.39                           & \cellcolor[rgb]{ .851,  .851,  .851}\textbf{0.44} \bigstrut[b]\\
\hline
\multirow{3}[2]{*}{Mean}       & ADD-S (\%) $\uparrow$          & 64.64                          & 92.39                          & 25.81                          & 22.29                          & 35.88                          & \cellcolor[rgb]{ .851,  .851,  .851}\textbf{96.52} \bigstrut[t]\\
                               & ADD (\%) $\uparrow$            & 33.36                          & 66.01                          & 16.54                          & 8.97                           & 16.08                          & \cellcolor[rgb]{ .851,  .851,  .851}\textbf{92.62} \\
                               & CD (cm) $\downarrow$           & 30.84                          & 52.05                          & 15.49                          & 52.57                          & 9.65                           & \cellcolor[rgb]{ .851,  .851,  .851}\textbf{0.57} \bigstrut[b]\\

\tableborder
\end{tabular}%
}
\vspace{-0.1in}
\caption{Per-video comparison on HO3D Dataset. ADD and ADD-S are AUC (0 to 0.1 $m$) percentage for pose evaluation. CD is the chamfer distance for shape reconstruction evaluation.}
\label{tab:ho3d_detail}
\vspace{-0.1in}
\end{table*}

\clearpage

%% file: tables/ycbineoat_detailed_quant.tex
\begin{table*}[h]
\centering
\def\mywidth{0.97\textwidth} 
\resizebox{\mywidth}{!}{
\begin{tabular}{c|c|ccccccccc}
\tableborder
Object                         & Metric                         & MaskFusion* \cite{runz2018maskfusion} & TEASER++* \cite{Yang20troteaser} & BundleTrack* \cite{bundle2021wen} & BundleTrack \cite{bundle2021wen} & \multicolumn{1}{c|}{DROID-SLAM \cite{teed2021droid}} & KinectFusion \cite{newcombe2011kinectfusion} & NICE-SLAM \cite{Zhu2022CVPR}   & SDF-2-SDF \cite{slavcheva2018sdf} & \cellcolor[rgb]{ .847,  .847,  .847}Ours \bigstrut\\
\hline
\multirow{3}[2]{*}{003\_cracker\_box} & ADD-S (\%) $\uparrow$          & 88.28                          & 81.35                          & 89.41                          & 90.20                          & 27.25                          & 56.04                          & 54.23                          & 19.89                          & \cellcolor[rgb]{ .847,  .847,  .847}\textbf{90.63} \bigstrut[t]\\
                               & ADD (\%) $\uparrow$            & 79.74                          & 63.24                          & 85.07                          & 85.08                          & 19.73                          & 42.73                          & 24.92                          & 12.13                          & \cellcolor[rgb]{ .847,  .847,  .847}\textbf{85.37} \\
                               & CD (cm) $\downarrow$           & -                              & -                              & -                              & 1.36                           & 2.95                           & 2.43                           & 4.03                           & 3.12                           & \cellcolor[rgb]{ .847,  .847,  .847}\textbf{0.76} \bigstrut[b]\\
\hline
\multirow{3}[2]{*}{021\_bleach\_cleanser} & ADD-S (\%) $\uparrow$          & 43.31                          & 82.45                          & 94.72                          & \textbf{95.22}                 & 27.13                          & 53.98                          & 17.96                          & 30.63                          & \cellcolor[rgb]{ .847,  .847,  .847}94.28 \bigstrut[t]\\
                               & ADD (\%) $\uparrow$            & 29.83                          & 61.83                          & 89.34                          & \textbf{89.34}                 & 12.83                          & 40.94                          & 9.55                           & 14.21                          & \cellcolor[rgb]{ .847,  .847,  .847}87.46 \\
                               & CD (cm) $\downarrow$           & -                              & -                              & -                              & 1.31                           & 2.43                           & 1.99                           & 9.40                           & 3.87                           & \cellcolor[rgb]{ .847,  .847,  .847}\textbf{0.53} \bigstrut[b]\\
\hline
\multirow{3}[2]{*}{004\_sugar\_box} & ADD-S (\%) $\uparrow$          & 45.62                          & 81.42                          & 90.22                          & 90.68                          & 53.87                          & 45.20                          & 14.56                          & 24.57                          & \cellcolor[rgb]{ .847,  .847,  .847}\textbf{93.81} \bigstrut[t]\\
                               & ADD (\%) $\uparrow$            & 36.18                          & 51.91                          & 85.56                          & 85.49                          & 43.38                          & 30.53                          & 8.70                           & 14.87                          & \cellcolor[rgb]{ .847,  .847,  .847}\textbf{88.62} \\
                               & CD (cm) $\downarrow$           & -                              & -                              & -                              & 2.25                           & 2.41                           & 2.56                           & 7.75                           & 1.70                           & \cellcolor[rgb]{ .847,  .847,  .847}\textbf{0.46} \bigstrut[b]\\
\hline
\multirow{3}[2]{*}{005\_tomato\_soup\_can} & ADD-S (\%) $\uparrow$          & 6.45                           & 71.61                          & 95.13                          & \textbf{95.24}                 & 0.08                           & 60.52                          & 17.08                          & 24.76                          & \cellcolor[rgb]{ .847,  .847,  .847}\textbf{95.24} \bigstrut[t]\\
                               & ADD (\%) $\uparrow$            & 5.65                           & 41.36                          & \textbf{86.00}                 & 85.78                          & 0.08                           & 45.64                          & 11.45                          & 10.89                          & \cellcolor[rgb]{ .847,  .847,  .847}83.10 \\
                               & CD (cm) $\downarrow$           & -                              & -                              & -                              & 7.36                           & \textbf{0.99}                  & 9.30                           & 1.52                           & 1.42                           & \cellcolor[rgb]{ .847,  .847,  .847}3.57 \bigstrut[b]\\
\hline
\multirow{3}[2]{*}{006\_mustard\_bottle} & ADD-S (\%) $\uparrow$          & 13.11                          & 88.53                          & 95.35                          & \textbf{95.84}                 & 42.29                          & 17.88                          & 8.77                           & 44.51                          & \cellcolor[rgb]{ .847,  .847,  .847}95.75 \bigstrut[t]\\
                               & ADD (\%) $\uparrow$            & 11.55                          & 71.92                          & \textbf{92.26}                 & 92.15                          & 15.10                          & 16.01                          & 7.33                           & 18.23                          & \cellcolor[rgb]{ .847,  .847,  .847}89.87 \\
                               & CD (cm) $\downarrow$           & -                              & -                              & -                              & 1.76                           & 2.90                           & 6.88                           & 7.95                           & 2.91                           & \cellcolor[rgb]{ .847,  .847,  .847}\textbf{0.45} \bigstrut[b]\\
\hline
\multirow{3}[2]{*}{Mean}       & ADD-S (\%) $\uparrow$          & 41.88                          & 81.17                          & 92.53                          & 93.01                          & 32.12                          & 46.39                          & 23.41                          & 28.20                          & \cellcolor[rgb]{ .847,  .847,  .847}\textbf{93.77} \bigstrut[t]\\
                               & ADD (\%) $\uparrow$            & 35.07                          & 57.91                          & \textbf{87.34}                 & 87.26                          & 20.39                          & 34.68                          & 12.70                          & 14.04                          & \cellcolor[rgb]{ .847,  .847,  .847}86.95 \\
                               & CD (cm) $\downarrow$           & -                              & -                              & -                              & 2.81                           & 2.34                           & 4.63                           & 6.13                           & 2.61                           & \cellcolor[rgb]{ .847,  .847,  .847}\textbf{1.16} \bigstrut[b]\\

\tableborder
\end{tabular}%
}
\vspace{-0.1in}
\caption{Per-object comparison (following the same protocol as \cite{bundle2021wen}) on YCBInEOAT Dataset. Results of MaskFusion*~\cite{runz2018maskfusion}, TEASER++*~\cite{Yang20troteaser} and BundleTrack*~\cite{bundle2021wen} are copied from the leaderboard in \cite{bundle2021wen}. For BundleTrack, we re-run the algorithm with the same segmentation masks as ours for fair comparison, and we augment with TSDF Fusion~\cite{curless1996volumetric,zeng20163dmatch} for reconstruction evaluation. ADD and ADD-S are AUC (0 to 0.1 $m$) percentage for pose evaluation. CD is the chamfer distance for shape reconstruction evaluation.}
\label{tab:ycbineoat_detail}
\vspace{-0.1in}
\end{table*}

%% file: tables/behave_detailed_quantA.tex
\clearpage
\begin{table*}[h]
\centering
\def\mywidth{0.9\textwidth} 
\resizebox{\mywidth}{!}{
\begin{tabular}{c|c|cccccc}
\tableborder
Video                          & Metric                         & DROID-SLAM \cite{teed2021droid} & BundleTrack \cite{bundle2021wen} & KinectFusion \cite{newcombe2011kinectfusion} & NICE-SLAM \cite{Zhu2022CVPR}   & SDF-2-SDF \cite{slavcheva2018sdf} & \cellcolor[rgb]{ .847,  .847,  .847}Ours \bigstrut\\
\hline
\multirow{3}[2]{*}{Date03\_Sub03\_boxlarge.2} & ADD-S (\%) $\uparrow$          & 72.59                          & 52.88                          & 21.09                          & 7.05                           & 24.78                          & \cellcolor[rgb]{ .847,  .847,  .847}\textbf{92.63} \bigstrut[t]\\
                               & ADD (\%) $\uparrow$            & 21.04                          & 13.00                          & 11.00                          & 3.02                           & 7.97                           & \cellcolor[rgb]{ .847,  .847,  .847}\textbf{86.72} \\
                               & CD (cm) $\downarrow$           & 8.61                           & 11.61                          & 8.80                           & 24.79                          & 41.97                          & \cellcolor[rgb]{ .847,  .847,  .847}\textbf{1.46} \bigstrut[b]\\
\hline
\multirow{3}[2]{*}{Date03\_Sub03\_boxlong.3} & ADD-S (\%) $\uparrow$          & 44.05                          & 27.77                          & 5.59                           & 10.21                          & 54.87                          & \cellcolor[rgb]{ .847,  .847,  .847}\textbf{77.0} \bigstrut[t]\\
                               & ADD (\%) $\uparrow$            & 14.06                          & 20.31                          & 1.83                           & 1.58                           & 13.75                          & \cellcolor[rgb]{ .847,  .847,  .847}\textbf{32.58} \\
\end{tabular}%
}
\vspace{-0.1in}
\caption{Per-video comparison on BEHAVE Dataset. ADD and ADD-S are AUC (0 to 0.5 $m$) percentage for pose evaluation.  CD is chamfer distance for shape reconstruction evaluation.  Table continues on the next page.  (This is part 1 of 4.)}
\label{tab:behave_detailA}
\vspace{-0.1in}
\end{table*}

\clearpage

%% file: tables/behave_detailed_quantB.tex
\begin{table*}[h]
\centering
\def\mywidth{0.9\textwidth} 
\resizebox{\mywidth}{!}{
\begin{tabular}{c|c|cccccc}
\tableborder
Video                          & Metric                         & DROID-SLAM \cite{teed2021droid} & BundleTrack \cite{bundle2021wen} & KinectFusion \cite{newcombe2011kinectfusion} & NICE-SLAM \cite{Zhu2022CVPR}   & SDF-2-SDF \cite{slavcheva2018sdf} & \cellcolor[rgb]{ .847,  .847,  .847}Ours \bigstrut\\
\hline
\multirow{3}[2]{*}{Date03\_Sub03\_tablesquare\_sit.3} & ADD-S (\%) $\uparrow$          & 81.23                          & 99.09                          & 98.97                          & 64.13                          & 57.54                          & \cellcolor[rgb]{ .847,  .847,  .847}\textbf{99.1} \bigstrut[t]\\
                               & ADD (\%) $\uparrow$            & 78.30                          & 98.65                          & 98.26                          & 33.85                          & 35.25                          & \cellcolor[rgb]{ .847,  .847,  .847}\textbf{98.71} \\
                               & CD (cm) $\downarrow$           & 3.04                           & 1.49                           & \textbf{1.13}                  & 37.66                          & 36.43                          & \cellcolor[rgb]{ .847,  .847,  .847}2.22 \bigstrut[b]\\
\hline
\multirow{3}[2]{*}{Date03\_Sub03\_toolbox.3} & ADD-S (\%) $\uparrow$          & 0.08                           & 26.69                          & 2.50                           & 5.96                           & 9.01                           & \cellcolor[rgb]{ .847,  .847,  .847}\textbf{92.39} \bigstrut[t]\\
                               & ADD (\%) $\uparrow$            & 0.08                           & 20.25                          & 1.44                           & 3.53                           & 1.52                           & \cellcolor[rgb]{ .847,  .847,  .847}\textbf{68.97} \\
\end{tabular}%
}
\vspace{-0.1in}
\caption{Per-video comparison on BEHAVE Dataset, continued from previous page.    (This is part 2 of 4.)}
\label{tab:behave_detailB}
\vspace{-0.1in}
\end{table*}

\clearpage

%% file: tables/behave_detailed_quantC.tex
\begin{table*}[h]
\centering
\def\mywidth{0.9\textwidth} 
\resizebox{\mywidth}{!}{
\begin{tabular}{c|c|cccccc}
\tableborder
Video                          & Metric                         & DROID-SLAM \cite{teed2021droid} & BundleTrack \cite{bundle2021wen} & KinectFusion \cite{newcombe2011kinectfusion} & NICE-SLAM \cite{Zhu2022CVPR}   & SDF-2-SDF \cite{slavcheva2018sdf} & \cellcolor[rgb]{ .847,  .847,  .847}Ours \bigstrut\\
\hline
\multirow{3}[2]{*}{Date03\_Sub04\_tablesmall\_lean.0} & ADD-S (\%) $\uparrow$          & 78.16                          & 98.44                          & 96.66                          & 17.29                          & 33.80                          & \cellcolor[rgb]{ .847,  .847,  .847}\textbf{98.49} \bigstrut[t]\\
                               & ADD (\%) $\uparrow$            & 66.19                          & 95.09                          & 87.52                          & 14.88                          & 18.06                          & \cellcolor[rgb]{ .847,  .847,  .847}\textbf{95.34} \\
                               & CD (cm) $\downarrow$           & 13.51                          & 8.27                           & \textbf{7.89}                  & 46.25                          & 40.48                          & \cellcolor[rgb]{ .847,  .847,  .847}9.36 \bigstrut[b]\\
\hline
\multirow{3}[2]{*}{Date03\_Sub04\_tablesmall\_lift.3} & ADD-S (\%) $\uparrow$          & \textbf{43.3}                  & 18.38                          & 10.62                          & 18.74                          & 37.41                          & \cellcolor[rgb]{ .847,  .847,  .847}30.33 \bigstrut[t]\\
                               & ADD (\%) $\uparrow$            & \textbf{26.87}                 & 8.81                           & 6.95                           & 7.78                           & 9.85                           & \cellcolor[rgb]{ .847,  .847,  .847}11.81 \\
\end{tabular}%
}
\vspace{-0.1in}
\caption{Per-video comparison on BEHAVE Dataset, continued from previous page.    (This is part 3 of 4.)}
\label{tab:behave_detailC}
\vspace{-0.1in}
\end{table*}

\clearpage

%% file: tables/behave_detailed_quantD.tex
\begin{table*}[ht]
\centering
\def\mywidth{0.9\textwidth} 
\resizebox{\mywidth}{!}{
\begin{tabular}{c|c|cccccc}
\tableborder
Video                          & Metric                         & DROID-SLAM \cite{teed2021droid} & BundleTrack \cite{bundle2021wen} & KinectFusion \cite{newcombe2011kinectfusion} & NICE-SLAM \cite{Zhu2022CVPR}   & SDF-2-SDF \cite{slavcheva2018sdf} & \cellcolor[rgb]{ .847,  .847,  .847}Ours \bigstrut\\
\hline
\multirow{3}[2]{*}{Date03\_Sub05\_toolbox.1} & ADD-S (\%) $\uparrow$          & 55.27                          & 24.17                          & 23.30                          & 13.54                          & 52.24                          & \cellcolor[rgb]{ .847,  .847,  .847}\textbf{89.64} \bigstrut[t]\\
                               & ADD (\%) $\uparrow$            & 36.92                          & 19.30                          & 15.41                          & 6.45                           & 29.98                          & \cellcolor[rgb]{ .847,  .847,  .847}\textbf{71.47} \\
                               & CD (cm) $\downarrow$           & 7.80                           & 17.49                          & 4.94                           & 38.37                          & 26.47                          & \cellcolor[rgb]{ .847,  .847,  .847}\textbf{2.94} \bigstrut[b]\\
\hline
\multirow{3}[2]{*}{Date03\_Sub05\_trashbin.3} & ADD-S (\%) $\uparrow$          & 78.78                          & 56.89                          & 24.44                          & 32.29                          & 40.09                          & \cellcolor[rgb]{ .847,  .847,  .847}\textbf{92.2} \bigstrut[t]\\
                               & ADD (\%) $\uparrow$            & 48.88                          & 16.38                          & 14.24                          & 14.19                          & 6.16                           & \cellcolor[rgb]{ .847,  .847,  .847}\textbf{56.67} \\
\end{tabular}%
}
\vspace{-0.1in}
\caption{Per-video comparison on BEHAVE Dataset, continued from previous page.    (This is part 4 of 4.)}
\label{tab:behave_detailD}
\vspace{-0.1in}
\end{table*}